%% file: main.tex
\documentclass[11pt]{article}

\usepackage[accent=orange]{pangram}
\usepackage[normalem]{ulem}
\usepackage{amsfonts}
\usepackage{amssymb}
\usepackage{graphicx}
\usepackage{amsmath}
\usepackage{algorithm}
\usepackage{algpseudocode}
\usepackage{tikz}
\usepackage{booktabs}
\usepackage{array}
\usepackage{longtable}
\usepackage{multirow}
\usepackage{float}
\usetikzlibrary{arrows.meta,backgrounds,calc,fit,positioning}
\definecolor{p4orange}{HTML}{E86A33}
\definecolor{p4blue}{HTML}{3A78C2}
\definecolor{p4human}{HTML}{2A9D8F}
\definecolor{p4assist}{HTML}{E9A23B}
\definecolor{p4ai}{HTML}{7656A7}
\definecolor{p4ink}{HTML}{263238}
\definecolor{p4muted}{HTML}{607D8B}

\tikzset{
  p4box/.style={
    draw=p4ink!70,
    rounded corners=2pt,
    fill=white,
    align=center,
    inner sep=5pt,
    minimum height=9mm
  },
  p4flow/.style={
    -{Latex[length=2.2mm]},
    thick,
    draw=p4ink!75
  },
  p4panel/.style={
    rounded corners=4pt,
    draw=p4ink!25,
    inner sep=8pt
  }
}

\title{Pangram 4 Technical Report}

\DeclareRobustCommand{\pangramaffilmark}{%
  \kern0.08em
  \raisebox{0.45ex}{\includegraphics[height=0.54em]{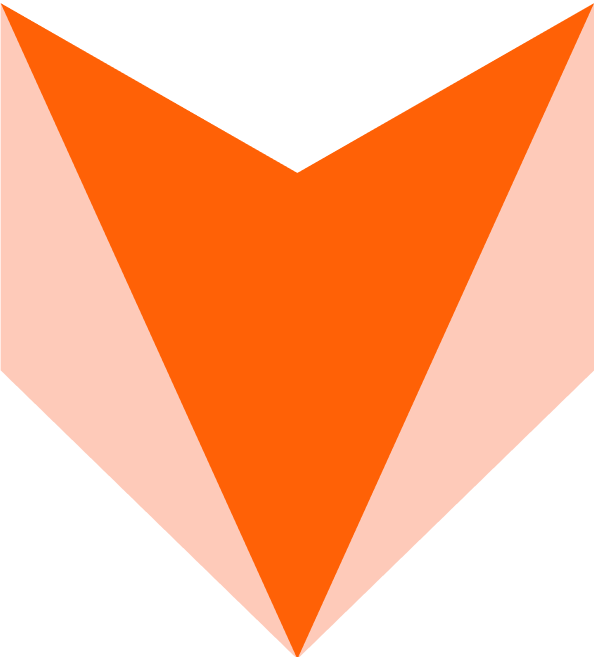}}}
\DeclareRobustCommand{\umdaffilmark}{%
  \kern0.08em
  \raisebox{0.18ex}{\includegraphics[height=0.82em]{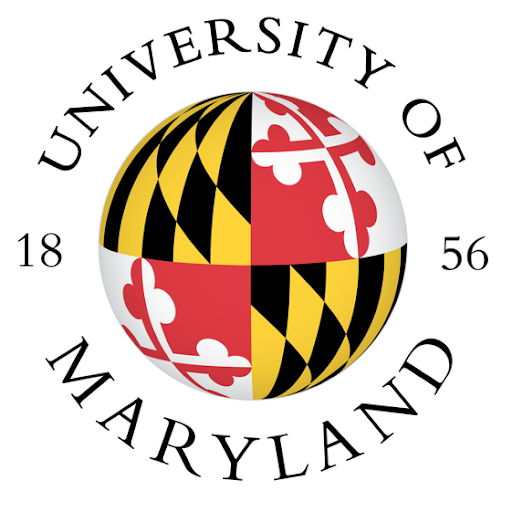}}}
\definecolor{codexblue}{HTML}{1F5F8B}

\author{%
  \begin{tabular}{@{}l@{}}
    {\normalfont\Authfont Ben    Glickenhaus\pangramaffilmark\textsuperscript{*}\enspace
      Katherine Thai\pangramaffilmark\textsuperscript{*}\enspace
      Jenna Russell\umdaffilmark}\enspace
      Elyas Masrour\pangramaffilmark\enspace
      Yue Han\pangramaffilmark\\
    {\normalfont\Authfont\kern-0.30em Max Spero\pangramaffilmark\enspace
      Bradley Emi\pangramaffilmark}\\
    {\normalfont\Affilfont
      \pangramaffilmark\,Pangram Labs\quad
      \umdaffilmark\,University of Maryland}\\
  \end{tabular}%
}

\pangramlogo{pangram-icon.png}
\pangramlogoheight{1.05cm}
\newcommand{\pfour}{\textbf{\color{pangram@accent}Pangram 4}\xspace}

\definecolor{darkorange}{HTML}{CC6600}

\definecolor{lightred}{HTML}{e99090}

\runningtitle{ Pangram 4 Technical Report}

\begin{document}
\maketitle
\begingroup
\renewcommand{\thefootnote}{\fnsymbol{footnote}}
\footnotetext[1]{Equal contribution.}
\endgroup

\input{sections/abstract}
\input{sections/1_intro}








\newpage
\tableofcontents
\newpage

\input{sections/2_data}
\input{sections/3_task}
\input{sections/4_training}
\input{sections/5_eval}
\input{sections/7_conclusion}









\bibliographystyle{plainnat}
\bibliography{references}

\appendix
\include{sections/appendix}

\end{document}

%% file: sections/abstract.tex
\begin{abstract}
\label{sec:abstract}

We present \pfour, the latest deep-learning-based AI-text classification model from Pangram Labs. \textbf{We achieve an AUROC of 0.9916 with a false positive rate of 0.0041\% and a false negative rate of 0.3396\%}.
In addition to its increased overall accuracy compared with Pangram 3, \pfour exhibits superior out-of-distribution generalization and robustness to adversarial attacks. Another novel contribution of \pfour is its improved ability to distinguish fine-grained edits and mixed AI-human co-authored text. We demonstrate improvements to both boundary detection tasks and the detection of interleaved AI assistance. Finally, we report metrics on standard AI detection benchmarks showing that \pfour achieves state-of-the-art performance on the AI text detection task across a wide variety of settings and domains.
\end{abstract}

%% file: sections/1_intro.tex
\section{Introduction}
\label{sec:intro}

The world is being inundated with an unprecedented volume of AI-generated text. Over a third of new internet material \cite{dolezal2026impactaigeneratedtextinternet}, 26\% of long-form social media posts \cite{pangram2026socialmedia}, 9\% of news articles \cite{russell2026ainews}, and 21\% of machine-learning conference reviews \cite{pangram2025iclr} are largely or fully AI-generated. ``AI slop,'' a term for unwanted AI content, is used to describe text that is obviously AI-generated \cite{shaib2026measuringaisloptext} or that offloads cognition onto the recipient.\footnote{\url{https://ampblog.substack.com/p/what-makes-slop-slop}}
AI is a tool with the power to automate many long-horizon tasks, but easily mis and overused with the potential for societal harm.

While it has been shown that individuals who regularly use AI can detect AI text by eye \cite{russell2025people}, this is a laborious task that requires both time and experience to do well. This creates an effort asymmetry between readers and writers: AI can produce well-formed text that could plausibly have come from an expert, nearly for free, but readers still incur the burden of understanding whether the content they are ingesting is meaningful, grounded, and well-researched. This strains existing systems that lack a way to reliably discern human from AI text.

Additionally, people are finding ways to incorporate AI into their writing workflows, producing documents of mixed authorship. These uses, while legitimate, do not diminish the societal harms caused by LLM-powered bots and undisclosed, fully AI-generated content polluting information ecosystems \cite{raj2026disclosurepenalty, drayson2025machine, yu2026retrievalcollapses}. To effectively mitigate these harms, it is imperative to understand the degree of human and AI involvement in written documents. This task motivates the creation of an AI-text detection model with high accuracy and granularity on AI, human, and AI-assisted text.

We introduce \pfour, the most capable AI text detection model developed by Pangram Labs. Designed to detect frontier models such as Claude Fable 5~\citep{anthropic2026fable} and GPT-5.6 Sol~\citep{openai2026gpt56}, \pfour achieves the highest accuracy on the AI-text detection task while maintaining a false positive rate of 0.0041\% (roughly 1 in 24,000). We build on Pangram 3, an AI detection model based on the EditLens architecture \citep{thai2026editlens}, with data and architectural improvements. \pfour represents a new frontier in AI text detection technology, with state-of-the-art precision and recall, performance on mixed-authorship text, fine-grained span predictions, and robustness to humanizer attacks. 

\begin{figure}[!t]
  \centering
  \includegraphics[width=\linewidth]{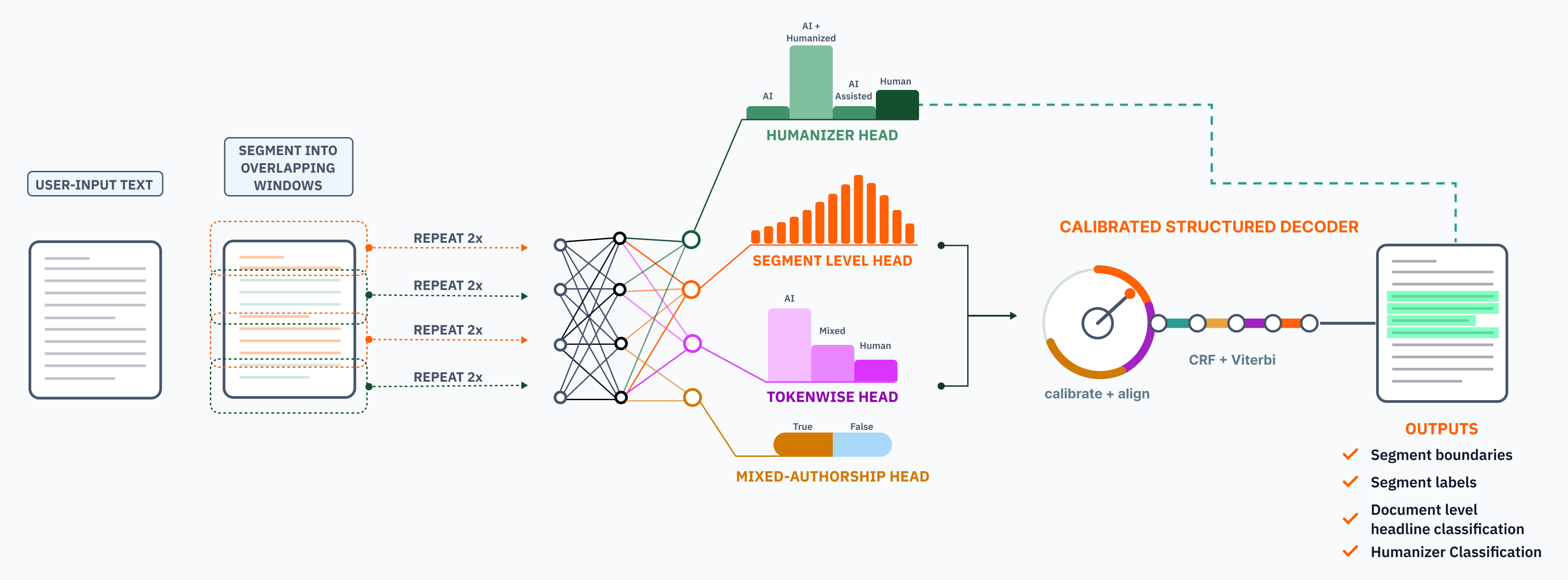}
  \caption{Overview of the \pfour architecture.}
  \label{fig:p4-overview}
\end{figure}

%% file: sections/2_data.tex
\section{Data}
\label{sec:data}

In this section, we report on the composition of our training set. We train on a wide distribution of domain sources, languages, and generator models. The human-written sources in the training dataset undergo various transformations, including synthetic mirroring, AI-editing, and translation, to produce the AI-generated and AI-edited texts that are present in the dataset. 

\subsection{Human-written Datasets}

\begin{figure}[!htbp]
    \centering
    \includegraphics[width=\textwidth]{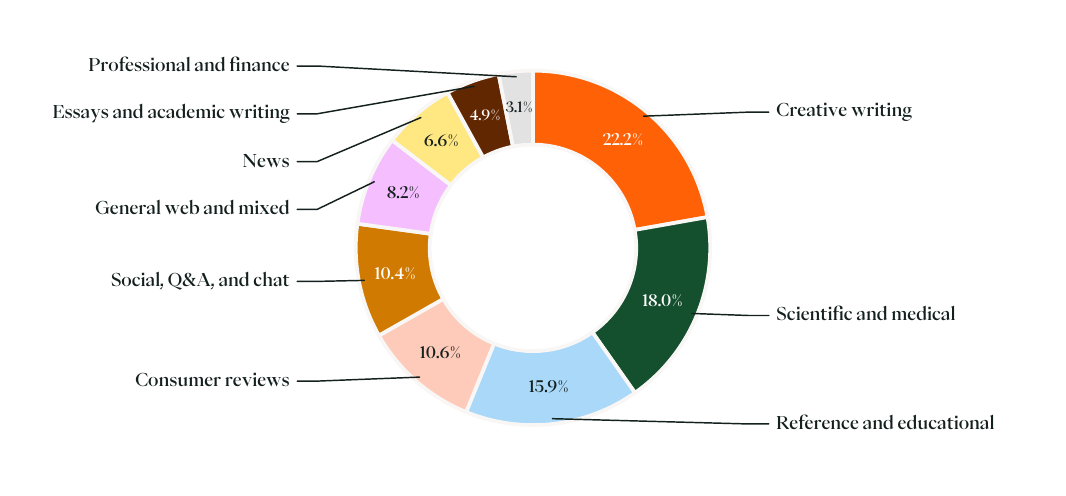}
    \caption{Approximate breakdown of human source text by category.}
    \label{fig:domains}
\end{figure}

Pangram is trained on a variety of datasets consisting of human writing from several sources, domains, and languages, similar to corpora used to train large language models. All datasets used to train Pangram are either openly licensed for commercial use or their rights are owned by the company. \textbf{\pfour is not trained on user-submitted data, customer data from API users, or data obtained by unauthorized Internet crawls}. A rough distribution of the categories of text used to train the model is presented in \autoref{fig:domains}.


\subsection{AI-Generated Datasets \& Synthetic Mirroring}

In keeping with previous Pangram models, the AI examples in our dataset are generated entirely in-house from a distribution of the most widely used LLMs. 
Our synthetic mirroring algorithm is described in the original Pangram technical report \cite{emi2024technicalreportpangramaigenerated}. This method is used to produce a diverse dataset of AI text and ensure topic invariance. We want the model to learn ``How was this text written?'' as opposed to ``What is this text about?'' An example of a synthetic mirror is as follows:

\paragraph{Synthetic mirror example}
The synthetic mirror is produced in two steps:

\begin{enumerate}
  \item \textbf{Prompt:} ``What is the topic of this article? $\langle$\textit{Original Document}$\rangle$'' \\
        \textbf{LLM Response:} topic $X$
  \item \textbf{Prompt:} ``Write an article about $X$'' \\
        \textbf{LLM Response:} the \emph{synthetic mirror} of the original document
\end{enumerate}

When available, synthetic mirrors are grounded by additional metadata such as article titles or keywords. For question-and-answer datasets, the question alone may suffice as a synthetic mirror prompt.
To prevent synthetic mirrors from repeating human text, a final check compares the generated text against its original source and discards examples where a significant portion of the original document is repeated verbatim.

\subsection{AI-Assisted Text}

Several recent works \cite{zhang2024mixset, saha2025apteval, artemova2025beemo, saha2026reviewpolishing, quaremba2026tsmbench, dycke2026youraitextmine,bsharat2026operationguidedprogressivehumantoaitext} benchmark AI detectors' ability to detect AI-polished, mixed, or otherwise co-authored writing that does not fall neatly into either the human-written or AI-generated categories. Our dataset consists of AI-assisted text in addition to human-written text and fully AI-generated text. We generate the AI-assisted text according to the method described in the EditLens paper \cite{thai2026editlens}.

%% file: sections/3_task.tex
\section{Task Formulation}
\label{sec:task}

\subsection{What is AI-generated Text?}

In order to define a tractable problem, we must first scope our definition of AI-generated text. We cannot define this as ``any text that was produced by any large language model.'' A large language model can simply memorize and reproduce human text, either from its training set or from its context window. Additionally, a large language model can also produce purely factual information, such as the answers to ``What is the capital of France?'' or ``What is Martin Luther King Jr.'s birthday?'' Answers to both of these questions would be ambiguous, and thus must be considered out of scope.

Pangram defines AI-generated text as \textbf{original, substantial natural-language prose generated by an LLM in response to an open-ended writing task or question}. Pangram is designed to detect responses to open-ended prompts and questions, rather than responses to questions that have a single correct answer. 

In addition to the open-generation criterion, we also require that, for AI text, the number of output tokens is greater than the number of input tokens. While AI summarization is a popular editing task, we must consider the output to be human if the AI does not add any original tokens of its own.
Finally, we require that text be at least 50 words long to be considered for analysis. This is a necessary condition for the model to have enough signal to make a confident prediction. This excludes short, conversational replies from Pangram's scope.

\subsection{Heterogeneous vs. Homogeneous Mixed Text}

\begin{figure}[htbp!]
    \centering
    \includegraphics[width=\textwidth]{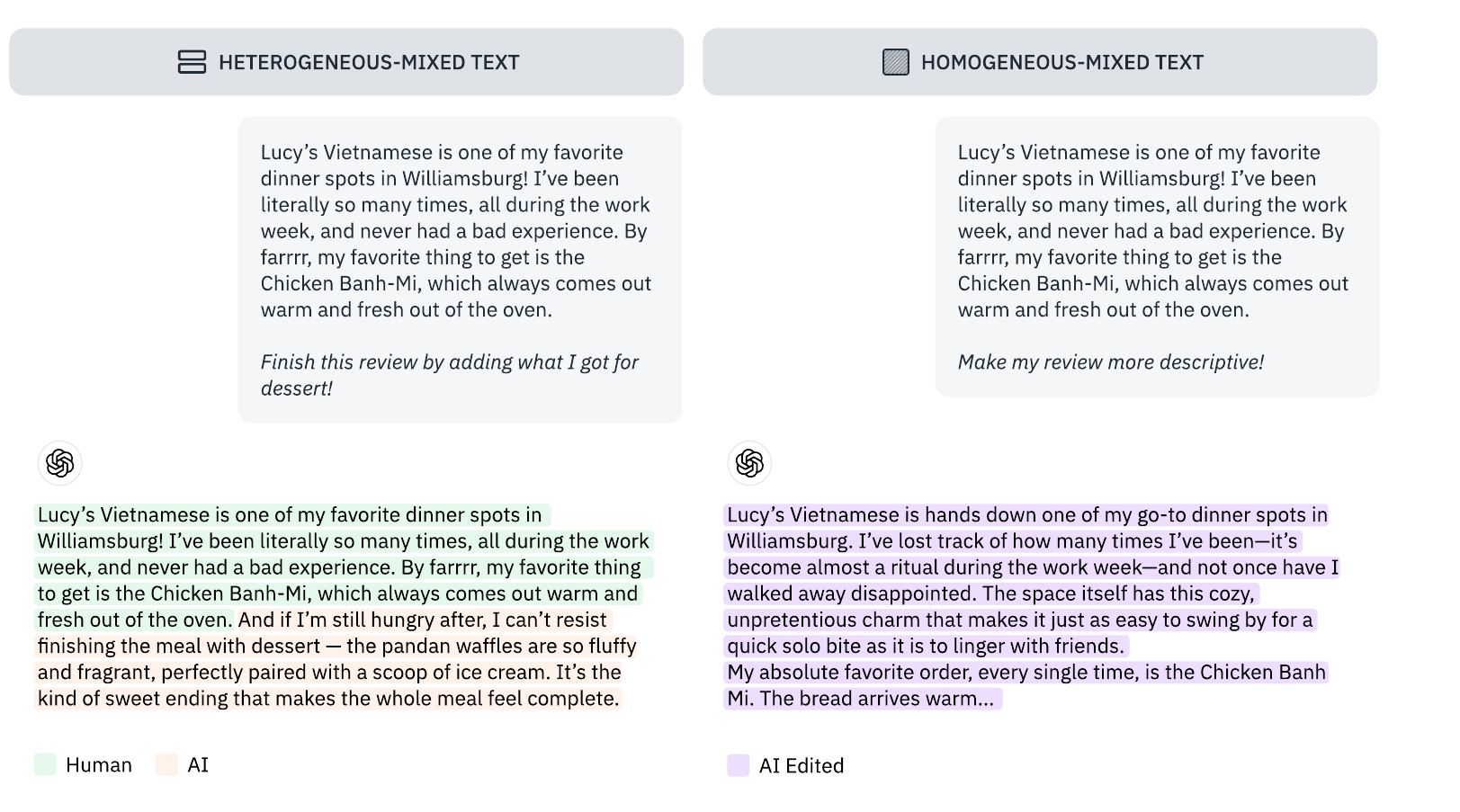}
    \caption{The difference between heterogeneous and homogeneous mixed text. In heterogeneous mixed text, each token has a well-defined author. In homogeneous mixed text, authorship attribution is mixed and ambiguous.}
    \label{fig:hetero_vs_homo}
\end{figure}

To motivate our labeling and architecture choices, we first introduce the concept of heterogeneous vs. homogeneous mixed text, which was originally introduced in the EditLens paper \cite{thai2026editlens}. In heterogeneous mixed text, each token has a clear authorship label, and each segment can be directly attributed to a single author, either human or AI. An example is a situation where a human writes a paragraph and then asks AI to continue it.

In homogeneous mixed text, authorship is entangled by the editing process. 
An example of this kind of mixed text is a human who writes a paragraph and then goes back and forth with an AI to edit it. In this situation, the paragraph no longer has a distinct author; there are contributions from both.

Additionally, the provenance of the surface style of a text can be distinct from the provenance of its underlying ideas \cite{russell2026storyscopeinvestigatingidiosyncrasiesai}. A human can have a very well-formed, concrete idea that is then worded and rephrased by an AI assistant: this is also an example of co-authorship. We use this to motivate our labeling schema, wherein we aim to differentiate fully AI-generated text from text where a human came up with the original idea, and then the AI was responsible for the final lexical choices that appear in the writing.

\subsection{Task Definition}

To our knowledge, \pfour is the first AI detection model that is able to identify both homogeneous and heterogeneous mixed writing \textit{simultaneously in a single pass.} \pfour takes an input sequence of $N$ tokens and produces an output of logits \(Z_{\mathrm{tok}} \in \mathbb{R}^{N \times 3}\) over the
classes \{\textsc{human}, \textsc{ai-assisted}, \textsc{ai-generated}\} for each token. 

\begin{enumerate}
    \item \textsc{human} means that the text was solely authored by humans, or that the amount of AI contribution is so low as to be negligible (e.g., light copyediting, literal translation, spelling and grammar fixes).
    \item \textsc{AI-Assisted} means that the text is co-authored \textbf{and} homogeneously mixed. 
    Although this label is nebulous for a single token, a segment of consecutive tokens with the \textsc{AI-Assisted} label indicates genuine co-authorship between humans and AI.
    \item \textsc{AI-Generated} means that the text is solely authored by AI, or that the text is significantly more AI-authored than human-authored---in this case, we would define text predicted as \textsc{AI-Generated} to be novel text from an LLM consistent with an open-ended prompt that allowed for original content solely from the LLM.
\end{enumerate}

Because these three labels are predicted on a per-token basis, a prediction of heterogeneous mixed text is expressed as a combination of these three labels across the sequence. We also note that, as a corollary of this formulation, text can be simultaneously heterogeneous and homogeneous mixed: there may be some sections that are co-authored, and some sections that have a single author.

\subsection{Humanization as an Auxiliary Task}
\label{subsec:humanization-task}
As Pangram becomes a more capable AI detector, adversarial actors have greater incentive to evade it. We define \emph{humanization} as the deliberate transformation or obfuscation of AI-generated text to evade AI-text detection \cite{masrour2025damage}. The auxiliary output estimates whether a document exhibits characteristics consistent with humanization.

We formulate humanization detection as a four-way, document-level classification task over {\textsc{human}, \textsc{ai-generated}, \textsc{humanized-ai}, \textsc{mixed-authorship}}. This supplemental classifier is applied when \pfour detects evidence of AI-generated content. It identifies evasion techniques such as deliberate typo injection, casing changes, synonym substitution, homoglyph attacks, and transformations produced by services that advertise humanization or AI-detector evasion. Incidental artifacts, such as PDF-extraction errors, copy-and-paste corruption, OCR errors, and encoding or Unicode-normalization problems, are not classified as humanization.
\subsection{Soft N-Grams Labeling}
\label{sec:softngrams}

We do not have ground-truth tokenwise labels for mixed text \textit{a priori}. Therefore, we compute tokenwise labels for AI-assisted text using the method described in Algorithm \autoref{alg:soft-ngram}, which we call Soft N-Grams Labeling. Following the assumptions in EditLens, we synthetically generate examples of mixed human- and AI-authored text by starting with confirmed human documents and applying AI edits to these texts. The input to the soft n-gram labeler is a human document $S$ and its corresponding AI-edited document $T$, which we refer to as the \textit{source} and the \textit{target}, respectively. 

\begin{algorithm}[!t]
\caption{Pangram soft n-gram labeling}
\label{alg:soft-ngram}
\begin{algorithmic}[1]
\Require Source text $S$, edited text $T$
\Ensure Labels for the edited text and AI fraction $f_{\mathrm{AI}}$

\State Split $S$ and $T$ into clauses
\ForAll{clauses $y$ in $T$}
    \State Find the most similar span $x$ in $S$
    \State Compute lexical similarity $L(x,y)$ using token and character n-grams
    \State Compute semantic similarity $E(x,y)$ using text embeddings

    \If{$L(x,y)$ is high}
        \State Label $y$ as \textsc{Human}
    \ElsIf{$L(x,y)$ or $E(x,y)$ indicates a substantial rewrite}
        \State Label $y$ as \textsc{AI-Assisted}
    \Else
        \State Label $y$ as \textsc{AI-Generated}
    \EndIf
\EndFor

\State Let $C_{\mathrm{H}}$, $C_{\mathrm{AA}}$, and $C_{\mathrm{AG}}$
       be the character counts for each label
\State $D \gets C_{\mathrm{H}}+C_{\mathrm{AA}}+C_{\mathrm{AG}}$
\State $f_{\mathrm{AI}} \gets
    \dfrac{0.5C_{\mathrm{AA}}+C_{\mathrm{AG}}}{D}$
\State \Return clause labels and $f_{\mathrm{AI}}$
\end{algorithmic}
\end{algorithm}

We diverge from EditLens, however, in the \textit{resolution} of the labeler. While EditLens (and Pangram 3) use a scalar distance metric in the embedding space of an LLM to compute the similarity between $S$ and $T$, in \pfour, we instead use a clause-level\footnote{A clause is defined as a group of words that contains a subject and a verb.} soft N-grams approach loosely inspired by the evaluation metric proposed in ROUGE \cite{ng-abrecht-2015-better}. 

\paragraph{Clause-level n-grams} The goal of the soft n-grams labeler is to extract granular provenance of each clause in the target $T$. We select the clause as the unit of the labeler after observing that the clause is the smallest atomic unit corresponding to a single expressible ``idea'' whose provenance can be traced. We begin by splitting the target $T$ into clauses using Claude Haiku 4.5 \cite{anthropic2025haiku45}. We also tried classical linguistic approaches to clause splitting, but found Claude Haiku more accurate and less brittle on unusual edge cases. For each clause in $T$, we ask the following questions:

\begin{enumerate}
    \item For this clause $y$, is there an exact or near-exact lexical match to this clause in $S$? If so, label this clause \textsc{Human}.
    \item If no lexical match is found, for this clause $y$, is there a clause in $S$ that semantically matches the idea in $y$? If so, label this clause \textsc{AI-Assisted}.
    \item Finally, if there is neither a lexical nor a semantic match to the clause $y$, then we deem the clause an original open-generation output and label this clause \textsc{AI-Generated}.
\end{enumerate}

This gives us tokenwise labels, where each token inherits the label from its parent clause. To convert the tokenwise labels into document labels, we define the weighted AI fraction $f_{AI}$, which is the fraction of characters labeled AI-generated, plus half the fraction of characters labeled AI-assisted. We then supervise this quantity as a bucketed regression target following the methodology in EditLens.

\paragraph{Deletion and rearrangement invariance}

We note that the Soft N-Grams Labeler is invariant to both deletions in the target $T$ and rearrangements of clauses between \(S\) and \(T\). We view this as a feature, not a bug: because at test time, our model only has access to $T$, it is ill-posed for the model to predict deletions from $S$, because it has no access to $S$. In informal language, it's impossible for the model to predict deletions because it can't see what was originally there. 
We also consider rearrangements to be non-edits from the perspective of the Soft N-Grams labeler. If all the sentences in a human-written document are merely reordered and not rewritten, we believe it is correct for our model to classify the text as human-written.

%% file: sections/4_training.tex
\section{Training}
\label{sec:training}

In this section, we describe the training procedure for \pfour. We first describe the model architecture, ablations run to pick the backbone, and individual task heads. We then describe the training process, active learning, and calibration to generate confidence estimates. 

\subsection{Architecture}

Here we describe the architecture of the model used to train token- and segment-level predictions, how calibration was conducted, and how we address humanized text. The model architecture is depicted in \autoref{fig:p4-overview}.

\paragraph{Backbone model}

The model architecture of \pfour is based on a popular open-weight MoE model. We attach a single dense layer as a custom classification head for each task that uses hidden states computed by the shared backbone at the final sequence position as inputs. For an input sequence of length \(S\), the shared backbone produces a hidden
representation \(\mathbf{h}_i \in \mathbb{R}^{D}\) at every position \(i\).
We attach an independent linear classification head for each task. We train on three window-level objectives: segment-level edits, mixed-authorship binary classification, and humanizer detection. These tasks use the representation at the final supervised
sequence position, \(\mathbf{h}_S\), as input. We use this approach (e.g., instead of mean pooling) because under causal attention,
\(\mathbf{h}_S\) has access to the complete input window. We also train a tokenwise
provenance head that instead applies its linear projection to every
\(\mathbf{h}_i\), producing a separate prediction at each token position.

\paragraph{Base model ablations}
We ran small-scale, iso-compute ablations against a variety of dense and sparse open-weight models to select the most promising candidates (performance is shown in \autoref{tab:backbone-ablations}).
For each candidate, we additionally swept attention, attention + dense, and attention + dense + routed expert (for MoEs) LoRA module targeting.
Each candidate is evaluated by its FNR at a fixed FPR on our holdout dataset.
All further experimentation is done on this base model. 

\begin{table}[!t]
    \centering
    \small
    \begin{tabular}{@{}llrrr@{}}
        \toprule
        Candidate
        & LoRA targets
        & Test FPR (\%) $\downarrow$
        & Test FNR (\%) $\downarrow$ \\
        \midrule
        A & Attention
        & 0.480 & 1.277 \\
        A & Attention + dense
           & 0.416 & \textbf{0.930} \\
        \addlinespace[2pt]

        B & Attention
          & 0.486 & 1.181 \\
        B & Attention + dense
          & 0.495 & 1.410 \\
        B & Attention + routed experts
           & 1.339 & 85.033 \\
        \addlinespace[2pt]

        C & Attention
          & 0.488 & 4.317 \\
        C & Attention + routed experts
           & 0.506 & 5.037 \\
        \addlinespace[2pt]

        D & Attention
           & 0.458 & 5.336 \\
        \bottomrule
    \end{tabular}
    \caption{
        Iso-compute backbone ablations. Backbone identities are anonymized.
    }
    \label{tab:backbone-ablations}
\end{table}

\paragraph{Segment predictions}

Following EditLens \cite{thai2026editlens}, we model the AI authorship attribution problem as a continuous spectrum rather than a binary classification task. Each labeled document is chunked into segments of length \(S = 512\) tokens. Each segment is then assigned a
weighted AI fraction
\[
    f_{AI}
    =
    \frac{0.5 C_{AA} + C_{AG}}
         {C_{H} + C_{AA} + C_{AG}},
\]
where \(C_{\textsc{H}}\), \(C_{\textsc{AA}}\), and
\(C_{\textsc{AG}}\) are character counts attributed to human,
AI-assisted, and AI-generated text, respectively. 
We discretize \(f_{AI} \in [0,1]\) into 15 ordered
buckets and train a segment-level classification head on the hidden state at
the final supervised sequence position.

\paragraph{Tokenwise predictions}
\pfour significantly increases prediction granularity. We are able to achieve this via a tokenwise prediction head. Let
\(H \in \mathbb{R}^{S \times D}\) denote the final-layer hidden states for a
window of \(S\) tokens with hidden dimension \(D\). A shared linear
token-classification head,
\[
    Z_{\mathrm{tok}} = H W_{\mathrm{tok}}^{\top},
    \qquad
    W_{\mathrm{tok}} \in \mathbb{R}^{3 \times D},
\]
produces logits \(Z_{\mathrm{tok}} \in \mathbb{R}^{S \times 3}\) over the
classes \{\textsc{human}, \textsc{ai-assisted}, \textsc{ai-generated}\} for
each token.

\paragraph{Repeat2} Tokenwise prediction introduces a new challenge with a causal-LM backbone. A segment-level classifier can use last-position logits to ensure the classification head attends to the entire sequence. Naively, this is inherently not possible for a tokenwise head---the hidden state for token \(x_i\) can ordinarily attend only to the prefix
\(x_{\leq i}\). This limitation manifests as significantly worse prediction accuracy on tokens at the start of a sequence. To fix this, we repeat the input sequence twice, as suggested by Repeat2 \cite{leviathan2025repeat2}, such that \[
    \widetilde{x} = (x_1,\ldots,x_S,x_1,\ldots,x_S).
\] This Repeat2 construction gives every supervised token access to the complete document context. We mask the token-classification loss over the entire first copy and supervise
only the corresponding tokens in the second copy:
\[
    \widetilde{y}
    =
    (\underbrace{\mathtt{ignore},\ldots,\mathtt{ignore}}_{S},
     y_1,\ldots,y_S).
\]

During training, Repeat2 is applied to each sampled tokenwise
window. At inference time, long documents are divided into overlapping windows
of at most \(S = 512\) tokens with a stride of \(256\) tokens, giving adjacent
full-length windows a \(50\%\) overlap. Repeat2 is applied to
each window, and the resulting predictions are aligned using their source-token
offsets and assembled before structured decoding. This allows us to aggregate logits for the same token across all windows in which it appears, such that the output is a document-length sequence with a three-class vector for each token in its original position.

\paragraph{Mixed-authorship head}
Stage~2 additionally trains a binary, window-level mixed-authorship head. Its
training target is positive when more than \(15\%\) of the supervised tokens
in a window belong to provenance classes other than the dominant class. We use this head to predict whether a window contains a meaningful mix of token labels, which is useful for downstream decoding as we will discuss in \cref{subsec:tokenwise-postprocessing}.

\paragraph{Humanizer Head}
\pfour introduces a new supplemental task: humanization detection. We train a probe on the shared last-position logits that learns to differentiate regular AI text from humanized AI text. We train this classification head as a probe to reduce the risk of humanizer-specific learned representations increasing our false positive rate on the main AI detection task. 
In controlled ablations, letting gradients flow through the base model from the humanizer head kept top-level FPR equivalent, but concentrated the failures in specific registers: academic writing and ESL essays. We opt to train the humanizer head as a probe to prevent the adverse impact of false positives.

\subsection{Training Process}
\pfour is trained using a two-stage process: first, we train a checkpoint supervised on just the segment and humanizer objectives. We then merge the LoRA adapters into the base model, initialize a fresh LoRA adapter, and continue training with the addition of the tokenwise objective. We use this two-stage approach as an efficiency optimization. The Repeat2 approach used in tokenwise training results in roughly half the throughput since we process twice as many tokens. Stage 1 training lets us bootstrap useful representations in a more compute-efficient manner, allowing us to train on the tokenwise objective for fewer total steps. \autoref{tab:p4-training-stages} summarizes the two stages.

\begin{table}[t]
    \centering
    \small
    \begin{tabular}{@{}p{0.08\linewidth}p{0.24\linewidth}
                         p{0.36\linewidth}p{0.23\linewidth}@{}}
        \toprule
        Stage & Initialization & Trained heads & Input and schedule \\
        \midrule
        1
        & Base backbone with a new LoRA adapter
        & Segment-level edit head (15-way); humanizer head
          (4-way, loss weight \(0.25\), stop-gradient)
        & Single-copy inputs of at most 512 tokens \\
        \addlinespace[3pt]
        2
        & Stage-1 adapter merged into the backbone; new LoRA adapter
        & Segment-level edit head (15-way); tokenwise provenance head
          (3-way); mixed-authorship head (2-way); humanizer head
          (4-way, loss weight \(0.25\), stop-gradient)
        & 512-token source windows using Repeat2 \\
        \bottomrule
    \end{tabular}
    \caption{
        \pfour training stages. Stop-gradient prevents the humanizer loss
        from updating the shared backbone in both stages.
    }
    \label{tab:p4-training-stages}
\end{table}

\paragraph{Active learning and hard negatives}
After training the \pfour model candidate, we do a round of offline active learning. This process involves running inference on a large pool of reserved data to find hard negatives: human text that the current checkpoint labels as AI. These samples are incorporated into our data pipeline, synthetically mirrored, and added to the training dataset. \pfour is then retrained from scratch on the resulting combined dataset.

\subsection{Tokenwise Postprocessing}
\label{subsec:tokenwise-postprocessing}

Compared to Pangram 3, \pfour uses a much more robust postprocessing algorithm that combines both token-level and segment-level signals and simultaneously handles calibration and smoothing. The new postprocessing algorithm is therefore much more precise in boundary detection and better calibrated on long documents.

\subsubsection{Motivation}

\paragraph{Long documents}
The model's context length is 512 tokens. In Pangram 3, postprocessing handles longer documents by repeatedly scoring candidate text spans, but this procedure is imprecise and computationally inefficient. Our new algorithm instead aggregates predictions over a single sliding-window pass and combines all observations in a CRF decoding problem.

\paragraph{Fine-grained Detection}
Boundary detection is particularly difficult for heterogeneous documents that
interleave human-written, AI-assisted, and AI-generated passages. A tokenwise
decoder increases the model's expressiveness by assigning one of these three
authorship states to every token before converting contiguous labels into
character-aligned segments.

\paragraph{Global Calibration}
In Pangram 3, we calibrated individual segments and subsequently aggregated
them. Consequently, the effect of aggregation on the
false positive and false negative tradeoff was difficult to control. The
tokenwise algorithm unifies calibration and postprocessing by producing a
calibrated unary score for every token and decoding all token labels jointly.

\subsubsection{Inputs and Outputs}

For a document containing $N$ source tokens, we divide the document into overlapping windows of at most 512 source tokens with a stride of 256; the final window is anchored to the end of the document. In Repeat2, the source tokens in each window are concatenated twice before model inference. Let $\mathcal{W}(i)$ denote the windows that contain source token $i$. Each window $w$ produces three relevant observations:
\begin{enumerate}
    \item a 15-bucket segment-head distribution indicating the overall estimate of the fraction of AI content in the segment:
          $\mathbf{p}^{\,s}_w \in \mathbb{R}^{15}$;
    \item a ternary token-head logit vector
          $\mathbf{z}^{\,t}_{w,i} \in \mathbb{R}^{3}$ for each token in the
          window; and
    \item a binary mixed-document distribution, whose positive-class
          probability is denoted $q_w$. We supervise this as an auxiliary task in which the model is trained to produce \textsc{True} if the text has any mix of AI and human content, and \textsc{False} if the text has homogeneous authorship.
\end{enumerate}

For a document, the post-processor returns a sequence of segments, each of which has a hard prediction of either \textsc{Human}, \textsc{AI-Assisted}, or \textsc{AI-Generated}.

\subsubsection{Algorithm}

\paragraph{Sliding-window inference}
We run a forward pass of the model for every overlapping window. This creates multiple model observations for tokens that appear in more than one window. We then collapse these observations into one ternary unary potential per source token and one transition score per adjacent token pair.

\paragraph{Segment-head projection}
The 15-bucket head estimates the weighted AI fraction as defined in \autoref{sec:softngrams}. Each bucket is projected onto three class anchors using triangular basis functions:
\begin{align}
    \phi_{\mathrm{human}}(f)
        &= \max(0,1-2f), \\
    \phi_{\mathrm{assisted}}(f)
        &= \max\!\left(0,1-2\lvert f-0.5\rvert\right), \\
    \phi_{\mathrm{AI}}(f)
        &= \max(0,2f-1).
\end{align}
The ternary document prior for window $w$ is a weighted average of the basis functions weighted by the predicted bucket probability vector.
\[
    \pi_w(c)
    =
    \frac{
        \sum_{b=0}^{14} p^{\,s}_w(b)\phi_c(f_b)
    }{
        \sum_{c'}
        \sum_{b=0}^{14} p^{\,s}_w(b)\phi_{c'}(f_b)
    }.
\]
This prior peaks at human for bucket 0, AI-assisted near the middle buckets, and AI-generated near bucket 14.

\paragraph{Calibrated unary potentials}

Raw token and segment logits are averaged across all windows containing token $i$, reducing the multiple observations of each token from sliding-window inference into a single 6-vector.

For each token, the calibration feature vector is
\[
    \mathbf{x}_i
    =
    \begin{bmatrix}
        \ell^{\,t}_i(\mathrm{human}) &
        \ell^{\,t}_i(\mathrm{assisted}) &
        \ell^{\,t}_i(\mathrm{AI}) &
        \ell^{\,s}_i(\mathrm{human}) &
        \ell^{\,s}_i(\mathrm{assisted}) &
        \ell^{\,s}_i(\mathrm{AI})
    \end{bmatrix}^{\!\top}.
\]
A multinomial logistic calibration model produces the CRF unary potential
\[
    u_i(c)
    =
    \alpha_c
    +
    \mathbf{w}_c^{\top}\mathbf{x}_i
    +
    \delta_c,
\]
where $\alpha_c$ is a fitted intercept, $\mathbf{w}_c$ contains the fitted
feature weights, and $\delta_c$ is an optional class-prior adjustment. These
parameters are fitted on held-out calibration data and may vary between model
releases.

\paragraph{Linear-chain CRF}
The calibrated unary scores define a three-state linear-chain conditional
random field. For adjacent labels $a$ and $b$, the Potts transition potential
at boundary $i$ is
\[
    \tau_i(a,b)
    =
    \begin{cases}
        0,
            & a=b, \\
        \min\!\left(0,-\lambda+\gamma m_i\right),
            & a\neq b,
    \end{cases}
\]
where $m_i$ is the center-weighted mixed log-odds at the boundary,
$\lambda\geq 0$ is the smoothness penalty, and $\gamma$ controls how strongly
mixed evidence can relax that penalty. Both values are selected using
calibration data.

The maximum a posteriori label sequence is
\[
    \widehat{\mathbf{T}}
    =
    \underset{\mathbf{T}\in\{0,1,2\}^{N}}{\arg\max}
    \left[
        \sum_{i=1}^{N}u_i(T_i)
        +
        \sum_{i=1}^{N-1}\tau_i(T_i,T_{i+1})
    \right].
\]


\paragraph{Decoding and marginals}
Multiclass Viterbi decoding computes $\widehat{\mathbf{T}}$. A numerically
stable forward--backward pass separately computes the token marginals
\[
    P(T_i=c\mid\text{all model observations}).
\]
Viterbi labels determine segment boundaries, while the marginals are retained
as token-level class probabilities.

\paragraph{Sentence majority voting}
Product output is constrained to sentence-level resolution. Sentences end at
terminal punctuation (including a trailing quote or bracket), a blank line, or
the end of the document. Tokens are assigned to sentences by character
midpoint. Every token in a sentence is replaced by the sentence's majority
Viterbi label. Deterministic tie-breaking is used to make the operation fully
reproducible.

\paragraph{Minimum segment length}
After sentence voting, the decoder repeatedly finds the first contiguous label
run shorter than a configured minimum $L_{\min}$ and merges it into a neighbor:
\begin{enumerate}
    \item if only one neighbor exists, that neighbor's label is used;
    \item if both neighbors have the same label, that label is used;
    \item otherwise, the longer neighboring run supplies the label; and
    \item a deterministic rule resolves equal-length ties.
\end{enumerate}
Merging continues until every remaining run has at least $L_{\min}$ tokens or
the document contains only one run. The value of $L_{\min}$ is a configurable
postprocessing choice rather than a fixed property of the model.

\subsubsection{Segments and Document-Level Prediction}

Adjacent tokens with the same final constrained label are coalesced and mapped
back to a complete partition of the original character sequence. Document
fractions are computed by character length:
\[
    f_c
    =
    \frac{
        \text{number of source characters assigned to class }c
    }{
        \text{number of source characters in the document}
    }.
\]

\subsubsection{Confidence}

Token class marginals are computed from the CRF before sentence-majority
voting and minimum-run merging. Token confidence is the largest marginal
probability, and segment confidence is the mean token confidence within the
segment. Therefore, confidence measures the peakedness of the unconstrained
CRF posterior; it is not a recalibrated probability that the final constrained
label is correct.

%% file: sections/5_eval.tex
\section{Evaluations}
\label{sec:eval}

We evaluate \pfour on human, AI-generated, and AI-assisted data over a range of in-domain and challenging benchmarks. The goal is to evaluate performance across representative real-world use cases across varying domains and languages.

\subsection{Evaluation Metrics}

Our final prediction $Z_{\mathrm{tok}} \in \mathbb{R}^{S \times 3}$ where $S$ is the number of segments as described in Section \ref{sec:training} must be decoded into a document-level prediction for evaluation purposes.

For purposes of evaluation, we decode the final tokenwise output into a documentwise output using the following rule-based method:

\[
\widehat{y} =
\begin{cases}
\mathrm{Human}, & \text{if } f_{\mathrm{human}} \ge 0.90,\\
\mathrm{AI},    & \text{if } f_{\mathrm{AI}} \ge 0.80,\\
\mathrm{Mixed}, & \text{otherwise}.
\end{cases}
\]

When we evaluate our model on fully human and fully AI text, we use standard binary classification metrics. We define a \textbf{False Positive} to be a prediction of \textsc{Mixed} or \textsc{AI} when the ground truth is \textsc{Human}, and a \textbf{False Negative} to be a prediction of \textsc{Human} or \textsc{Mixed} when the ground truth is AI. Note that a prediction of \textsc{Mixed} counts as a failure for both false positive rate and false negative rate measurements, making our metrics stricter than threshold-based calibration on a binary logit distribution as is typical for other methods.
For AUROC, TPR @ X\% FPR and other rank-based metrics, we directly use the post-processed weighted AI fraction to rank the texts in the corpus.

\subsection{Overall False Negative Rate}
We generate 520,000 AI outputs and run Pangram on them to identify false negatives. At the production operating point, \pfour achieves an \textbf{overall false negative rate of 0.3396\%}. This is an improvement over Pangram 3.3.2, which reports a FNR of 1.9942\% on the same dataset.

\paragraph{Synthetic benchmark construction}
We evaluate \pfour on a comprehensive synthetic benchmark of 520,000 examples generated by 26 frontier open-source and closed-weight
models. First, we use \texttt{Mistral-Small-24B-Instruct-2501} \cite{mistralsmall3} to filter prompts from the Chatbot Arena Conversations Dataset~\cite{lmsys_chatbot_arena}. We chose the Chatbot Arena Conversations dataset so that the prompts in the benchmark reflect the real-world distribution of user prompts to LLMs. We filtered for prompts that request an open-ended generation or pose an open-ended question. Multiple-choice questions, math, and coding-related inquiries were considered out of scope for this benchmark. After identifying 20,000 in-scope prompts, we generate a response for each prompt using each of the 26 LLMs for a total of 520,000 attempted samples, of which 519,993 were
successfully scored. We found 1,766 false negatives, yielding an FNR of 0.3396\%, with a 95\% Wilson confidence interval of (0.3242\%, 0.3558\%).

\paragraph{FNR by generation model} \autoref{tab:fnr-by-generator} displays results by generator model. Anthropic models have the lowest FNR, at 0.195\%, and OpenAI models have an FNR of 0.316\%. 
We find that there is no significant correlation between release date and FNR.

\begin{table}[!t]
  \centering
  \scriptsize
  \setlength{\tabcolsep}{2pt}
  \renewcommand{\arraystretch}{0.95}
  \resizebox{\linewidth}{!}{%
  \begin{tabular}{@{}llllrr@{}}
    \toprule
    Company & Model family & Generator & Release date & Pangram 3.3.2 & \pfour \\
    \midrule
    Anthropic & Claude
      & Claude Fable 5 \citep{anthropic2026fable} & Jun. 9, 2026 & 1.030\% & 0.255\% \\
      & & Claude Opus 5 \citep{anthropic2026opus5} &  Jul. 24, 2026 & 0.585\% & 0.195\% \\
      & & Claude Sonnet 5 \citep{anthropic2026sonnet5} & Jun. 30, 2026 & 0.750\% & 0.140\% \\
      & & Claude Opus 4.8 \citep{anthropic2026opus48}& May 28, 2026 & 0.885\% & 0.235\% \\
      & & Claude Sonnet 4.6 \citep{anthropic2026sonnet46} & Feb.~17, 2026 & 1.955\% & 0.215\% \\
      & & Claude Haiku 4.5 \cite{anthropic2025haiku45} & Oct.~15, 2025 & 0.805\% & 0.130\% \\
      & & \textit{Family aggregate}
      & -- & \textbf{1.002\%} & \textbf{0.195\%} \\
      \midrule
    \addlinespace[0.25em]
    DeepSeek & V4
      & DeepSeek V4 Flash \citep{deepseek2026v4} & Apr.~24, 2026 & 1.625\% & 0.455\% \\
      & & DeepSeek V4 Pro \citep{deepseek2026v4} & Apr.~24, 2026 & 2.050\% & 0.320\% \\
      & & \textit{Family aggregate}
      & -- & \textbf{1.838\%} & \textbf{0.388\%} \\
       \midrule
    \addlinespace[0.25em]
    Google & Gemini
      & Gemini 3.1 Pro Preview \citep{google2026gemini31} & Feb.~19, 2026 & 6.380\% & 0.555\% \\
      & & Gemini 3.5 Flash \citep{google2026gemini35} & May~19, 2026 & 3.895\% & 0.440\% \\
      & & \textit{Family aggregate}
      & -- & \textbf{5.138\%} & \textbf{0.498\%} \\
       \midrule
    \addlinespace[0.25em]
    Google & Gemma
      & Gemma 4 31B IT \citep{google2026gemma4} & Apr.~2, 2026 & 2.910\% & 0.430\% \\
       \midrule
    \addlinespace[0.25em]
    Meta & Llama
      & Llama 3.3 70B Instruct \cite{meta2024llama33} & Dec.~6, 2024 & 4.405\% & 0.550\% \\
       \midrule
    \addlinespace[0.25em]
    Mistral AI & Mistral
      & Mistral Medium 3.5 \citep{mistral2026medium35} & Apr.~29, 2026 & 0.835\% & 0.400\% \\
       \midrule
    \addlinespace[0.25em]
    Moonshot AI & Kimi
      & Kimi K2.6 \citep{moonshot2026k26} & Apr.~20, 2026 & 2.560\% & 0.395\% \\
       \midrule
    \addlinespace[0.25em]
    NVIDIA & Nemotron
      & Nemotron 3 Ultra 550B-A55B \citep{nvidia2026nemotron3ultra} & Jun.~4, 2026 & 1.320\% & 0.310\% \\
       \midrule
    \addlinespace[0.25em]
    OpenAI & GPT
      & GPT-5.6 Sol \citep{openai2026gpt56} & Jul.~9, 2026 & 3.240\% & 0.450\% \\
      & & GPT-5.6 Terra \citep{openai2026gpt56} & Jul.~9, 2026 & 2.040\% & 0.345\% \\
      & & GPT-5.5 \citep{openai2026gpt55} & Apr.~23, 2026 & 1.645\% & 0.345\% \\
      & & GPT-5.4 \citep{openai2026gpt54} & Mar.~5, 2026 & 1.145\% & 0.270\% \\
      & & GPT-5.4 Mini \citep{openai2026gpt54} & Mar.~17, 2026 & 0.900\% & 0.330\% \\
      & & GPT-OSS 120B \citep{openai2025gptoss} & Aug.~5, 2025 & 0.395\% & 0.155\% \\
      & & \textit{Family aggregate}
      & -- & \textbf{1.561\%} & \textbf{0.316\%} \\
       \midrule
    \addlinespace[0.25em]
    Alibaba Cloud & Qwen
      & Qwen 3.7 Max \citep{qwen2026modelreleases} & May~19, 2026 & 2.545\% & 0.315\% \\
       \midrule
    \addlinespace[0.25em]
    Tencent & Hunyuan
      & HY3 Preview \citep{tencent2026hy3preview} & Apr.~23, 2026 & 1.830\% & 0.330\% \\
       \midrule
    \addlinespace[0.25em]
    Thinking Machines & Inkling
      & Inkling \citep{thinkingmachines2026inkling} & Jul.~15, 2026 & 0.300\% & 0.190\% \\
       \midrule
    \addlinespace[0.25em]
    xAI & Grok
      & Grok 4.3 \citep{xai2026models} & Apr.~17, 2026 & 2.645\% & 0.605\% \\
       \midrule
    \addlinespace[0.25em]
    Z.ai & GLM
      & GLM 5.2 \citep{zai2026glm52} & Jun.~13, 2026 & 3.175\% & 0.470\% \\
    \midrule
    \multicolumn{4}{@{}l}{\textbf{Overall}}
      & \textbf{1.9942\%} & \textbf{0.3396\%} \\
    \bottomrule
  \end{tabular}
  }
  \caption{False negative rate by generator model and family.}
  \label{tab:fnr-by-generator}
  \vspace{0.35em}
\end{table}


\subsection{Overall False Positive Rate}
\label{subsec:fpr}
We evaluate \pfour on over a million human-written English texts and find that we have a false positive rate of 0.0041\%, with a 95\% confidence interval of (0.0032\%, 0.0050\%). To put this in perspective, this is one false positive for every $\sim$24,300 requests. This is also an improvement to Pangram 3.3.2, which has a false positive rate of 0.0539\% on the same dataset.

\subsection{Performance by Length}
\begin{figure}[H]
  \centering
  \begin{minipage}[t]{0.49\linewidth}
    \centering
    \includegraphics[width=\linewidth]{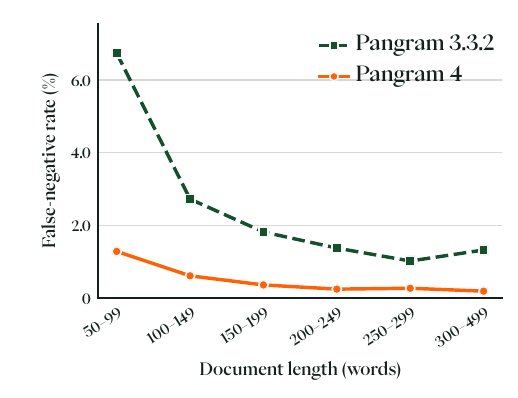}
  \end{minipage}\hfill
  \begin{minipage}[t]{0.49\linewidth}
    \centering
    \includegraphics[width=\linewidth]{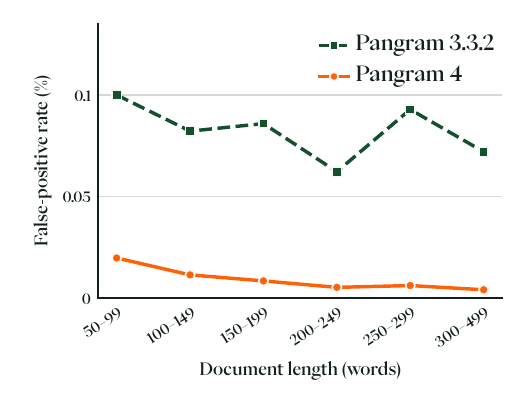}
  \end{minipage}
  \caption{Error rates by document length for Pangram 3.3.2 and \pfour.}
  \label{fig:performance-by-length}
\end{figure}

AI detection is often harder on shorter texts, due to less text signal to compute predictions on. We evaluate performance by length, finding that \pfour outperforms Pangram 3.3.2 at every length bucket, as displayed in \autoref{fig:performance-by-length}.

\subsection{Performance on Mixed Authorship Text}

\subsubsection{Homogeneous Mixed Eval}

\paragraph{FPR on AI-Polished Text}
We define \textbf{AI-Assisted False Positive Rate (FPR)} as the rate at which a model incorrectly flags minor edits made by AI as fully AI-generated. To evaluate the performance of our model at distinguishing lightly AI-edited texts from fully AI-generated texts, we create a dataset of AI-proofread and AI-polished texts. We use consumer products (Grammarly\footnote{\url{https://app.grammarly.com/}}, Apple Intelligence, and Gemini in Google Docs) and frontier LLMs (Opus 4.8, Gemini 3.1, and GPT-5.5) to apply light edits to 11,363 pieces of human-written student writing from ELLIPSE \cite{ellipse}, PELIC \cite{pelic}, and ICNALE \cite{icnale}. For example, we used the Writing Tools feature embedded in Apple's Pages word editor to ``Proofread" some texts or prompted Opus 4.8 to ``Correct minor academic style issues while keeping the draft recognizably human." For the full list of editing tools and prompts, see Table \ref{app:polish-subset}. In Table \ref{tab:polish-prediction-distribution}, we show that Pangram 3.3.2 classifies AI-polished text as \textsc{AI} 0.18\% of the time. \pfour only flags AI-polished text as \textsc{AI} 0.01\% of the time---an 18x reduction in AI-Assisted FPR.

\begin{table}[H]
\centering
\footnotesize
\renewcommand{\arraystretch}{1.25}
\begin{tabular}{@{}lrrr@{}}
\toprule
\textbf{Model}
& \textbf{Human}
& \textbf{Mixed}
& \textbf{AI (FPR)} \\
\midrule
Pangram 3.3.2
& 11,329 (99.70\%)
& 14 (0.12\%)
& 20 (0.18\%) \\
\pfour
& 11,301 (99.45\%)
& 61 (0.54\%)
& \textbf{1 (0.01\%)} \\
\bottomrule
\end{tabular}
\caption{Prediction distribution and AI-Assisted FPR on our AI polish evaluation set. Cells report count (percentage). $n$ = 11,363 texts.}
\label{tab:polish-prediction-distribution}
\vspace{0.5em}
\end{table}

\paragraph{Recall on AI-Edited Texts}
We define \textbf{AI-Assisted Recall} as the rate at which a model classifies an AI-assisted text as \textsc{Mixed}. Because there is no single definition for when a text should be considered AI-edited, we create two datasets with different standards for AI assistance.

We create the first dataset in a similar fashion to our AI Polish evaluation dataset. We use the same consumer products and frontier LLMs to apply editing instructions or prompts that demonstrate a clear intent to substantially modify the source text. For the full list of editing tools and prompts, see \autoref{app:edits-subset}. As in our AI Polish dataset, these edits are applied to 14,990 human-written student essays. In \autoref{tab:edits-prediction-distribution}, we show that Pangram 3.3.2 has an AI-Assisted Recall of only 5.54\%, predicting \textsc{Human} the majority (79.23\%) of the time. \pfour has a much-improved AI-Assisted Recall of 55.01\% on the same edited texts, predicting \textsc{Mixed} 10 times as often as Pangram 3.3.2.

\begin{table*}[h]
\centering
\footnotesize
\renewcommand{\arraystretch}{1.25}
\begin{tabular}{@{}lrrr@{}}
\toprule
\textbf{Model}
& \textbf{Human}
& \textbf{Mixed (Recall)}
& \textbf{AI} \\
\midrule
Pangram 3.3.2
& 11,877 (79.23\%)
& 831 (5.54\%)
& 2,282 (15.22\%) \\
\pfour
& 6,201 (41.37\%)
& \bfseries 8,246 (55.01\%)
& 543 (3.62\%) \\
\bottomrule
\end{tabular}
\caption{Prediction distribution and AI-Assisted Recall on our AI Editing Prompts evaluation set. Cells report count (percentage). $n$ = 14,990 texts.}
\label{tab:edits-prediction-distribution}
\end{table*}

We also create a second dataset where AI-assistance is defined by a quantitative
metric: the Soft N-grams Labeling approach from
\autoref{sec:softngrams}. First, we mine the WildChat dataset
\cite{zhao2024wildchat} using GPT-5.5 nano for over 5,000 unique editing prompts
submitted by users that are neither domain-specific (i.e., related to a field or
discipline) nor source-specific (i.e., mentioning specific words, phrases, or
arguments that should be modified). Then we randomly apply these edits using one
of Opus 4.8, Gemini 3.1, and GPT-5.5 to human-authored Reddit posts, student essays,
and creative writing. Finally, we score all resulting edited texts using Soft
N-Grams Labeling and filter our dataset to 4,826 examples with an AI fraction of
between 0.25 and 0.75. We deem texts with this score to be AI-assisted by any
reasonable standard, as a score of 0.25 means that half of the words have been
modified by AI or 25\% of the words have been newly generated by AI (or some
combination of the two). Conversely, a score of 0.75 indicates that the LLM
generated 75\% of the words in the human-written source. In table
\ref{tab:wildchat-prediction-distribution}, we show that Pangram 3.3.2 only
classifies 12.6\% of these AI-assisted texts as \textsc{Mixed}. Additionally,
Pangram 3.3.2 displays polarized behavior, preferring to classify these
AI-assisted texts as \textsc{Human} or \textsc{AI} rather than \textsc{Mixed}.
On the other hand, \pfour classifies the majority of these texts as
\textsc{Mixed}, achieving an AI-Assisted Recall of 65.7\%, five times that of Pangram
3.3.2.

\begin{table*}[h]
\centering
\footnotesize
\renewcommand{\arraystretch}{1.25}
\begin{tabular}{@{}lrrr@{}}
\toprule
\textbf{Model}
& \textbf{Human}
& \textbf{Mixed (Recall)}
& \textbf{AI} \\
\midrule
Pangram 3.3.2
& 3,150 (65.27\%)
& 609 (12.62\%)
& 1,067 (22.11\%) \\
\pfour
& 1,344 (27.85\%)
& \bfseries 3,145 (65.17\%)
& 337 (6.98\%) \\
\bottomrule
\end{tabular}
\caption{Prediction distribution and AI-Assisted Recall on our WildChat Edits evaluation set. Cells report count (percentage). $n$ = 4,826 texts.}
\label{tab:wildchat-prediction-distribution}
\end{table*}

\subsubsection{Heterogeneous Mixed Eval}

\paragraph{Dataset Construction}

For heterogeneous mixed eval, we aim to evaluate the resolution of our detector. We create datasets with interleaved human and AI text. These datasets contain documents that alternate $N$ fully human sentences with $N$ fully AI-generated sentences, for $N \in \{1, 4, 8, 12, 16, 20\}.$ We create these documents by taking long human-written documents from our test set, and replacing alternating $N$ sentence chunks with AI sentences from a random choice of GPT-5.4, Claude Sonnet 4.6, and Gemini 2.5 Flash. We sample the original human documents from our creative writing and formal academic writing splits.

\paragraph{Results}

In table \ref{tab:pangram3-vs-pangram4-resolution}, we report the tokenwise binary classification metrics stratified by $N$ on this heterogeneous mixed dataset, as well as the document mean average error (MAE), which is the average difference between the predicted fraction of AI tokens and the ground-truth fraction of AI tokens (lower is better). \pfour is significantly more accurate than Pangram 3.3.2 at all resolutions on these challenging interleaved texts.

Qualitatively, \pfour is able to predict AI text at a much finer-grained resolution than Pangram 3.3.2. We display a representative example below in Figure \ref{fig:heteromixed}.

\begin{figure}[!t]
    \centering
    \includegraphics[width=0.8\textwidth]{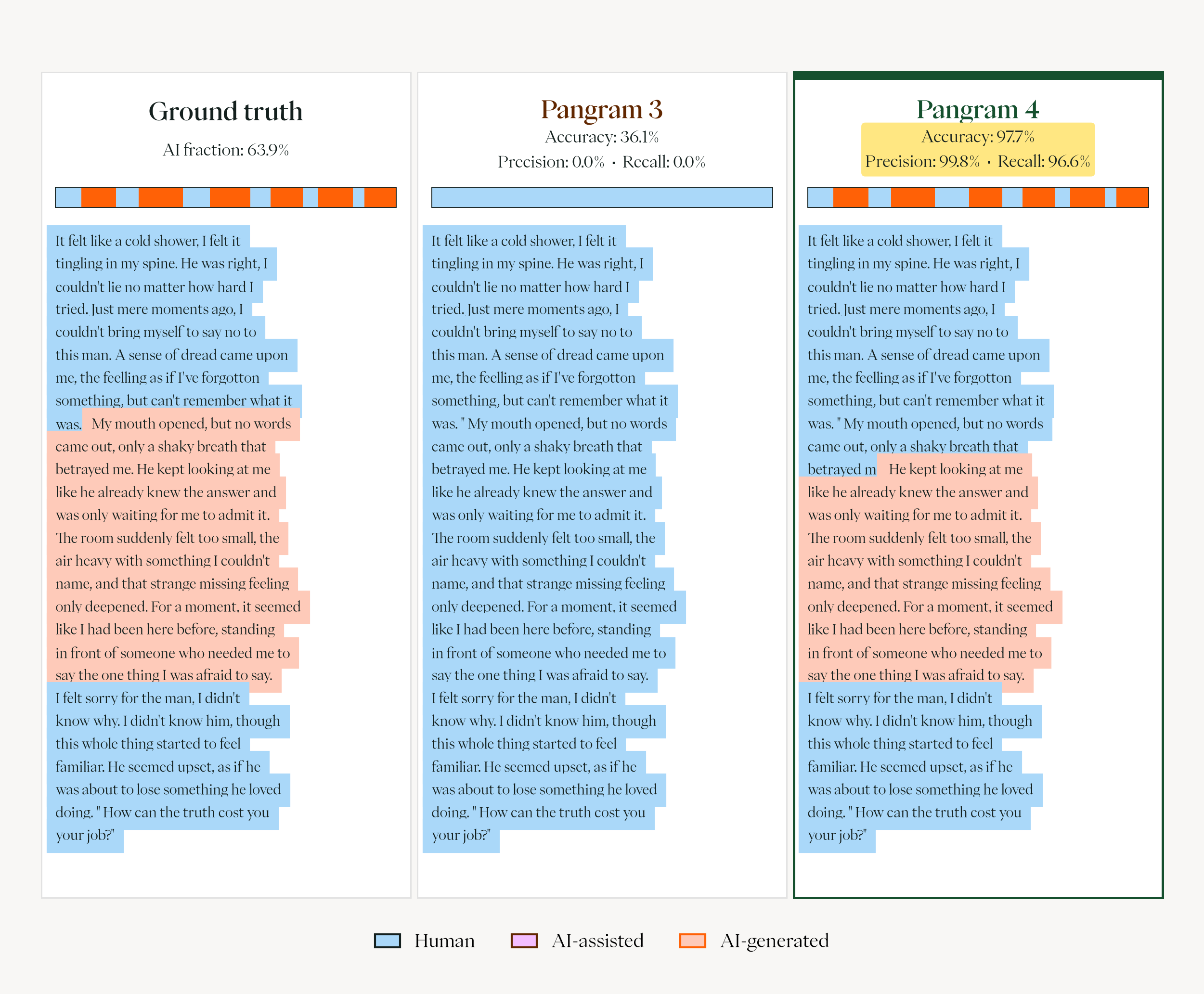}
    \caption{Pangram 3.3.2 vs. 4 on an excerpt from a heterogeneous mixed document. \pfour is able to detect AI at a much finer-grained} resolution.
    \label{fig:heteromixed}
\end{figure}

\begin{table}[H]
\centering
\small
\begin{tabular}{rlrrrr}
\toprule
Resolution & Model
& Tokenwise Acc. $\uparrow$
& Tokenwise Prec. $\uparrow$
& Tokenwise Recall $\uparrow$
& Doc MAE $\downarrow$ \\
\midrule

1
& Pangram 3.3.2 & 46.12\% & 63.25\% & 22.43\% & 45.01\% \\
& \textbf{\pfour}
& \textbf{75.02\%}
& \textbf{92.82\%}
& \textbf{62.86\%}
& \textbf{18.27\%} \\
\midrule

4
& Pangram 3.3.2 & 51.59\% & 56.88\% & 19.27\% & 40.60\% \\
& \textbf{\pfour}
& \textbf{92.41\%}
& \textbf{99.56\%}
& \textbf{85.43\%}
& \textbf{6.80\%} \\
\midrule

8
& Pangram 3.3.2 & 55.08\% & 56.21\% & 49.42\% & 31.80\% \\
& \textbf{\pfour}
& \textbf{95.94\%}
& \textbf{99.61\%}
& \textbf{92.31\%}
& \textbf{3.51\%} \\
\midrule

12
& Pangram 3.3.2 & 68.31\% & 65.67\% & 71.30\% & 25.39\% \\
& \textbf{\pfour}
& \textbf{97.12\%}
& \textbf{99.66\%}
& \textbf{94.33\%}
& \textbf{2.58\%} \\
\midrule

16
& Pangram 3.3.2 & 81.89\% & 76.35\% & 85.85\% & 16.68\% \\
& \textbf{\pfour}
& \textbf{97.50\%}
& \textbf{99.63\%}
& \textbf{94.73\%}
& \textbf{2.20\%} \\
\midrule

20
& Pangram 3.3.2 & 88.17\% & 82.02\% & 93.37\% & 11.76\% \\
& \textbf{\pfour}
& \textbf{98.18\%}
& \textbf{99.51\%}
& \textbf{96.31\%}
& \textbf{1.52\%} \\
\midrule

Overall
& Pangram 3.3.2 & 64.18\% & 68.33\% & 52.61\% & 28.60\% \\
& \textbf{\pfour}
& \textbf{92.02\%}
& \textbf{98.35\%}
& \textbf{85.45\%}
& \textbf{5.87\%} \\

\bottomrule
\end{tabular}
\caption{Comparison of Pangram 3.3.2 with adaptive boundaries and \pfour.
Bold values indicate the better result. Precision and recall treat AI text
as the positive class.}
\label{tab:pangram3-vs-pangram4-resolution}
\end{table}

\subsection{Performance on Languages Other than English}

\autoref{tab:error-rates-by-language} reports the FPR and FNR for supported languages. FPRs are measured on pre-2022 per-language subsets of FineWeb2 \cite{penedo2025fineweb2}. While not reported, we test all 104 languages reported in FineWeb2, finding only 14 false positives among 996,273 examples, for an \textbf{overall multilingual FPR of 0.0014\%} with a 95\% Wilson confidence interval of (0.0008\%, 0.0024\%). FNRs are measured across 18 supported languages using a test set of 190,149 multilingual AI generations, which are synthetic mirrors of English data, including how-to articles, news articles, and reviews. \pfour's \textbf{overall multilingual FNR is 1.24\%} with a 95\% Wilson confidence interval of (1.19\%, 1.29\%). We hypothesize that higher false negative rates in languages such as Urdu (5.31\%) and Persian (3.17\%) may originate from differences in the base model's tokenization; however, the effects of tokenization on multilingual AI-text detection remain unknown.

\begin{table}[h]
  \centering
  \footnotesize
  \setlength{\tabcolsep}{4pt}
  \begin{tabular}{@{}lrr@{\hspace{3em}}lrr@{}}
    \toprule
    Language & FPR & FNR & Language & FPR & FNR \\
    \midrule
    Arabic     & 0.0000\% & 0.9764\% & Persian    & 0.0000\% & 3.1715\% \\
    Chinese    & 0.0000\% & 0.7133\% & Polish     & 0.0000\% & ---      \\
    Czech      & 0.0000\% & 0.2760\% & Portuguese & 0.0078\% & 0.9946\% \\
    Dutch      & 0.0026\% & 0.8397\% & Romanian   & 0.0000\% & ---      \\
    French     & 0.0026\% & 1.7839\% & Russian    & 0.0052\% & 0.4850\% \\
    German     & 0.0026\% & 1.3258\% & Spanish    & 0.0026\% & 0.8867\% \\
    Greek      & 0.0000\% & ---      & Swedish    & 0.0000\% & ---      \\
    Hindi      & 0.0000\% & 1.3682\% & Turkish    & 0.0000\% & 1.1066\% \\
    Hungarian  & 0.0000\% & ---      & Ukrainian  & 0.0361\% & 1.5214\% \\
    Italian    & 0.0000\% & 0.3436\% & Urdu       & 0.0000\% & 5.3169\% \\
    Japanese   & 0.0000\% & 0.9305\% & Vietnamese & 0.0026\% & 1.9653\% \\
    Korean     & 0.0000\% & 0.5805\% &            &          &          \\
    \bottomrule
  \end{tabular}
  \caption{Multilingual false positive and false negative rates by language.}
  \label{tab:error-rates-by-language}
\end{table}

\subsection{Performance on Non-native English}
\paragraph{Datasets} 
There has been speculation and anxiety that AI detectors are biased against non-native English speakers \cite{liang2023bias}. To address this concern,  we evaluate \pfour's false positive rate on several public corpora of non-native English text: ELLIPSE \cite{ellipse}, ICNALE \cite{icnale} (a dialogue dataset), PELIC \cite{pelic}, and the TOEFL dataset by \citet{liang2023bias}. In \cref{tab:ell-fpr}, we show that \pfour has a single raw false positive out of 24,586 samples, resulting in a 0.0041\% FPR that is the same as our overall FPR reported in Section \ref{subsec:fpr}.

\begin{table}[h]
\centering
\small
\begin{tabular}{@{}lrrrrr@{}}
\toprule
& &
\multicolumn{2}{c}{\pfour} &
\multicolumn{2}{c}{Pangram 3.3.2} \\
\cmidrule(lr){3-4}
\cmidrule(l){5-6}
Dataset
& $N$
& FP
& FPR (\%)
& FP
& FPR (\%) \\
\midrule
ELLIPSE
& 3{,}899 & 0 & 0.000\% & 0 & 0.000\% \\

ICNALE
& 5{,}593 & 0 & 0.000\% & 0 & 0.000\% \\

PELIC
& 15{,}005 & 1 & 0.0067\% & 2 & 0.0133\% \\

Liang TOEFL
& 89 & 0 & 0.000\% & 0 & 0.000\% \\
\midrule
\textbf{Overall}
& \textbf{24{,}586}
& \textbf{1}
& \textbf{0.0041\%}
& \textbf{2}
& \textbf{0.0081\%} \\
\bottomrule
\end{tabular}
\caption{False positive rates on human-authored English learner writing and speech transcripts.}
\label{tab:ell-fpr}
\end{table}

\subsection{Public Benchmarks}

\input{sections/5_public_benchmark_summary}

\paragraph{UChicago}
\citet{jabarian2025artificial} pairs authentic human writing from several
genres with outputs from multiple LLMs across a range of document lengths. It
also includes
dedicated evaluations of ultra-short passages and texts processed by
commercial humanizers. Full detector comparisons and default-threshold
false positive rates are reported in \autoref{tab:uchicago-detectionai} and
\autoref{tab:uchicago-default-fpr}.

\paragraph{VUB}
\citet{vanvlasselaer2026whowrote} focuses on long-form academic papers and
evaluates fully human-written, fully AI-generated, hybrid, and humanized AI-generated
documents. Because the synthetic documents have known ground-truth AI
proportions, the benchmark can test whether detector scores reflect the extent
of AI involvement rather than merely provide binary classifications. Our
evaluation uses the public fully AI-generated subset, whose score distribution
is reported in \autoref{tab:vub-fully-ai}.

\paragraph{GEDE}
\citet{gehring2025gede} is an education-focused benchmark containing more than 900
student-written essays and over 12,500 LLM-generated essays. Its defining
feature is a spectrum of student-contribution levels, ranging from fully human
writing through light LLM improvement and full generation to adversarial
humanization. Results for every evaluated split are reported in
\autoref{tab:gede-results}.

\paragraph{Perkins et al}
\citet{perkins2024simple} evaluates 15 original AI-generated samples and 89
variants created using six detector-evasion techniques. It emphasizes
adversarial robustness and the risks that unreliable detection poses to
fairness and inclusivity in academic assessment. Baseline and manipulated-text
results are reported in \autoref{tab:perkins-results}.

\paragraph{Epoch AI Style Imitation}
\citet{epoch2026aidetectorsfalsenegatives} contains 495 human passages and 594
AI passages: 297
generated from basic prompts and 297 prompted to imitate the styles of sampled
authors. We report document-level false negative rates across both AI
conditions, span-level false negative rates for the Pangram models, and
false positive rates on the human controls in
\autoref{tab:detector-error-rates}.

\paragraph{DetectRL}
\citet{wu2024detectrl} evaluates detector robustness and generalization across domains,
generators, attack strategies, text lengths, and real-world factors in human
writing. Task~1 measures in-domain performance across domains, generators, and
attack types. Task~4 applies paraphrasing, perturbation, and data mixing to
human-written texts to test how these transformations affect detector
behavior. The full zero-shot and supervised leaderboard is reported in
\autoref{tab:detectrl_leaderboard}.

\paragraph{MELD-eval}
\citet{li2026meld} introduces a controlled generator-shift benchmark built from four
current-generation chat models and eight English domains. It contains
paired human texts, clean AI outputs, and attack-augmented AI outputs, and it
tests zero-shot transfer to held-out generators at low false positive rates.
Generator-level and attack-level results are reported in
\autoref{tab:meld_eval_by_generator} and
\autoref{tab:meld_eval_by_attack}, respectively.

\paragraph{Sem-Detect}
\citet{duarte2026semdetect} introduces a dataset of more than 20,000 human-written,
AI-generated, and LLM-refined peer reviews from ICLR and NeurIPS, together with
a detector that combines textual features with claim-level semantic analysis.
The benchmark tests whether detectors can distinguish fully AI-generated
reviews from authentic or LLM-refined human reviews whose original judgments
are preserved. Binary detection results are reported in
\autoref{tab:semdetect_binary}.

\paragraph{Saha Peer Review}
\citet{saha2026reviewpolishing} simulates several levels of human--AI
collaboration, including fully AI-generated reviews, AI-generated reviews
conditioned on human-provided key points, AI-polished human reviews, and
unmodified human reviews. Its easy and hard subsets vary the diversity of
generating models and prompts. The full cohort-level comparison is reported in
\autoref{tab:saha_peer_review}.

\paragraph{OpAI-Bench}
\citet{bsharat2026operationguidedprogressivehumantoaitext} constructs nine
successive versions of each human-written document
using five AI editing operations and increasing predefined levels of AI
coverage. Because it provides character-level authorship provenance, we compare
the predicted AI+Assisted fraction directly with the ground-truth AI character
fraction throughout the editing trajectory. Stage-level results are reported
in \autoref{tab:opai-progressive-edits}.

\newcommand{\appendixUChicagoResults}{%
\begin{table}[H]
\centering
\begin{tabular}{llrrr}
\toprule
Split & Detector & $N$ & AUROC & TPR @ 1\% FPR \\
\midrule
Standard, full length
    & \textbf{\pfour} & 15,936 & \textbf{0.999997} & \textbf{100.00\%} \\
    & Pangram 3          & 15,928 & 0.999884 & 99.86\% \\
    & Originality        & 14,613 & 0.998038 & 94.15\% \\
    & GPTZero            & 15,934 & 0.989643 & 98.64\% \\
\midrule
Standard, under 50 words
    & \textbf{\pfour} & 1,312 & \textbf{0.999888} & \textbf{99.70\%} \\
    & Pangram 3          & 1,311 & 0.995829 & 88.87\% \\
    & GPTZero            & 1,312 & 0.940549 & 0.00\% \\
\midrule
Humanizer, full length
    & \textbf{\pfour} & 15,932 & \textbf{0.999579} & \textbf{98.93\%} \\
    & Pangram 3          & 15,928 & 0.998566 & 98.08\% \\
    & Originality        & 10,812 & 0.962496 & 28.49\% \\
    & GPTZero            & 11,689 & 0.718367 & 44.32\% \\
\midrule
Humanizer, under 50 words
    & \textbf{\pfour} & 1,312 & 0.981019 & \textbf{73.32\%} \\
    & Pangram 3          & 1,312 & \textbf{0.983775} & 61.74\% \\
    & GPTZero            & 1,312 & 0.644036 & 0.00\% \\
\bottomrule
\end{tabular}
\caption{Performance on the UChicago DetectionAI benchmark. TPR is measured at a 1\% false positive rate.}
\label{tab:uchicago-detectionai}
\medskip
\begin{minipage}{0.96\textwidth}
\footnotesize
\textit{Note:} Originality is excluded from both under-50-word splits because it does not support texts of that length. $N$ is the number of observations for which each detector returned a usable score. GPTZero and Originality sometimes rejected the requests.
\end{minipage}
\end{table}

\begin{table}[H]
\centering
\begin{tabular}{lrrr}
\toprule
Detector & Human $N$ & False positives & FPR \\
\midrule
\textbf{\pfour} & 7,968 & \textbf{0} & \textbf{0.00\%} \\
Pangram 3          & 7,964 & 0          & 0.00\% \\
Originality        & 7,308 & 8          & 0.11\% \\
GPTZero            & 7,968 & 57         & 0.72\% \\
\bottomrule
\end{tabular}
\caption{Overall false positive rates on the UChicago DetectionAI human controls at each detector's default 50\% decision threshold.}
\label{tab:uchicago-default-fpr}

\medskip
\begin{minipage}{0.96\textwidth}
\footnotesize
\textit{Note:} A document is classified as AI-generated when its detector
score is at least 50\%.
\end{minipage}
\end{table}
}

\newcommand{\appendixVUBResults}{%
\begin{table}[H]
\centering
\begin{tabular}{lrrrrrr}
\toprule
Detector & $N$ & 80--100\% & 60--80\% & 40--60\% & 20--40\% & 0--20\% \\
\midrule
\textbf{\pfour}
    & 39 & \textbf{39} & \textbf{0} & \textbf{0} & \textbf{0} & \textbf{0} \\
Pangram 3
    & 39 & 25 & 13 & 0 & 1 & 0 \\
GPTZero
    & 39 & 0 & 0 & 0 & 12 & 27 \\
Copyleaks
    & 39 & 0 & 0 & 0 & 9 & 30 \\
Turnitin
    & 39 & 0 & 0 & 0 & 0 & 39 \\
\bottomrule
\end{tabular}
\caption{Score distribution on the public fully AI-generated subset of the VUB
benchmark.}
\label{tab:vub-fully-ai}
\end{table}
}

\newcommand{\appendixGEDEResults}{%

\begin{table}[H]
\centering
\scriptsize
\renewcommand{\arraystretch}{0.88}
\begin{tabular}{llrrr}
\toprule
Split & Detector & $N$ & AUROC & TPR @ 1\% FPR \\
\midrule

\multirow{8}{*}{Overall}
 & \pfour        & 569 & \textbf{1.000} & \textbf{100.0\%} \\
 & Pangram 3        & 569 & 0.999 & 98.9\% \\
 & GPTZero          & 569 & 0.979 & 77.6\% \\
 & RoBERTa          & 569 & 0.937 & 28.9\% \\
 & DetectGPT        & 542 & 0.917 & 33.7\% \\
 & Fast-DetectGPT   & 569 & 0.911 & 55.5\% \\
 & Ghostbuster      & 569 & 0.834 & 28.9\% \\
 & Intrinsic-Dim    & 569 & 0.554 & 0.0\% \\
\midrule

\multirow{8}{*}{Fully generated}
 & \pfour        & 285 & \textbf{1.000} & \textbf{100.0\%} \\
 & Pangram 3        & 285 & \textbf{1.000} & \textbf{100.0\%} \\
 & GPTZero          & 285 & 0.997 & 93.2\% \\
 & RoBERTa          & 285 & 0.994 & 54.7\% \\
 & Fast-DetectGPT   & 285 & 0.985 & 82.1\% \\
 & DetectGPT        & 269 & 0.984 & 69.1\% \\
 & Ghostbuster      & 285 & 0.960 & 52.1\% \\
 & Intrinsic-Dim    & 285 & 0.541 & 0.0\% \\
\midrule

\multirow{8}{*}{AI-improved human}
 & \pfour        & 285 & \textbf{1.000} & \textbf{100.0\%} \\
 & Pangram 3        & 285 & 0.999 & 97.4\% \\
 & GPTZero          & 285 & 0.953 & 54.7\% \\
 & RoBERTa          & 285 & 0.888 & 11.1\% \\
 & DetectGPT        & 274 & 0.847 & 6.6\% \\
 & Fast-DetectGPT   & 285 & 0.797 & 10.5\% \\
 & Ghostbuster      & 285 & 0.680 & 7.4\% \\
 & Intrinsic-Dim    & 285 & 0.509 & 0.0\% \\
\midrule

\multirow{8}{*}{Humanized}
 & \pfour        & 189 & \textbf{1.000} & \textbf{100.0\%} \\
 & Pangram 3        & 189 & \textbf{1.000} & \textbf{100.0\%} \\
 & GPTZero          & 189 & 0.995 & 92.6\% \\
 & Fast-DetectGPT   & 189 & 0.993 & 92.6\% \\
 & DetectGPT        & 181 & 0.928 & 18.9\% \\
 & RoBERTa          & 189 & 0.918 & 12.8\% \\
 & Ghostbuster      & 189 & 0.893 & 25.5\% \\
 & Intrinsic-Dim    & 189 & 0.671 & 0.0\% \\
\bottomrule
\end{tabular}
\caption{Performance on the GEDE public benchmark. TPR is measured at a
maximum empirical false positive rate of 1\%.}
\label{tab:gede-results}
\end{table}
}

\newcommand{\appendixPerkinsResults}{%
\begin{table}[H]
\centering
\begin{tabular}{lrrrr}
\toprule
& \multicolumn{2}{c}{Baseline AI}
& \multicolumn{2}{c}{Manipulated AI} \\
\cmidrule(lr){2-3}\cmidrule(lr){4-5}
Detector & $N$ & Mean accuracy & $N$ & Mean accuracy \\
\midrule
\textbf{\pfour} & 15 & \textbf{100.0\%} & 89 & \textbf{94.1\%} \\
Copyleaks & 15 & 73.9\% & 90 & 58.7\% \\
GPTZero   & 15 & 26.4\% & 90 & 16.7\% \\
Turnitin  & 15 & 50.0\% & 90 & 7.9\% \\
\bottomrule
\end{tabular}
\caption{Performance on the Perkins benchmark using the paper's
three-method mean accuracy.}
\label{tab:perkins-results}
\end{table}
}

\newcommand{\appendixDetectRLResults}{%
\begin{table}[H]
\centering
\scriptsize
\setlength{\tabcolsep}{2.2pt}
\renewcommand{\arraystretch}{1.08}
\resizebox{\textwidth}{!}{%
\begin{tabular}{l c cc cc cc ccc cc cc c}
\toprule
& &
\multicolumn{2}{c}{\textbf{Multi-Domain}}
& \multicolumn{2}{c}{\textbf{Multi-LLM}}
& \multicolumn{2}{c}{\textbf{Multi-Attack}}
& \multicolumn{3}{c}{\textbf{Generalization}}
& \multicolumn{2}{c}{\textbf{Time}}
& \multicolumn{2}{c}{\textbf{Human Writing}}
& \textbf{Avg.} \\
\cmidrule(lr){3-4}
\cmidrule(lr){5-6}
\cmidrule(lr){7-8}
\cmidrule(lr){9-11}
\cmidrule(lr){12-13}
\cmidrule(lr){14-15}
\textbf{Detector}
& \textbf{Evaluation Type}
& \textbf{AUROC} & \textbf{F1}
& \textbf{AUROC} & \textbf{F1}
& \textbf{AUROC} & \textbf{F1}
& \textbf{Domain F1}
& \textbf{LLM F1}
& \textbf{Attack F1}
& \textbf{Train F1}
& \textbf{Test F1}
& \textbf{AUROC}
& \textbf{F1}
& \textbf{F1} \\
\midrule

\textbf{\pfour}
& Zero-shot
& 97.02 & 96.93
& 97.32 & 97.25
& 97.22 & 96.00
& 96.95
& 97.25
& 95.99
& N/A
& 91.30
& 93.80
& 90.77
& 95.30\textsuperscript{\dag} \\

Pangram 3.3.2
& Zero-shot
& 99.75 & 98.05
& 99.85 & 98.39
& 99.72 & 97.53
& \textbf{98.05}
& \textbf{98.39}
& \textbf{97.53}
& N/A
& \textbf{93.73}
& 95.67
& 91.95
& \textbf{96.70\textsuperscript{\dag}} \\

Open Pangram
& Zero-shot
& 99.48 & 94.20
& 99.48 & 94.30
& 99.26 & 94.14
& 94.20
& 94.30
& 94.14
& N/A
& 88.10
& 96.44
& 91.60
& 93.12\textsuperscript{\dag} \\

\midrule

Rob-Base
& Supervised
& \textbf{99.98} & \textbf{99.75}
& \textbf{99.93} & \textbf{99.58}
& 99.56 & 97.66
& 83.00 & 91.81 & 92.37
& 79.99 & 74.00
& \textbf{97.34} & 94.31
& 93.02 \\

Rob-Large
& Supervised
& 99.78 & 98.87
& 95.16 & 90.03
& \textbf{99.87} & \textbf{99.03}
& 77.20 & 82.85 & 83.96
& \textbf{86.08} & 85.23
& 96.68 & \textbf{94.63}
& 91.49 \\

X-Rob-Base
& Supervised
& 99.92 & 99.34
& 99.14 & 98.17
& 98.49 & 96.07
& 75.97 & 92.73 & 90.58
& 84.25 & 73.83
& 93.43 & 90.29
& 91.71 \\

X-Rob-Large
& Supervised
& 99.01 & 97.44
& 97.40 & 93.47
& 99.31 & 97.75
& 76.14 & 85.89 & 73.42
& 86.35 & 79.83
& 97.21 & 94.43
& 90.59 \\

\midrule

Binoculars
& Zero-shot
& 83.95 & 78.25
& 83.30 & 74.83
& 85.05 & 78.53
& 77.47 & 74.10 & 74.70
& 73.82 & 74.34
& 90.68 & 85.98
& 79.61 \\

Revise-Detect.
& Zero-shot
& 67.24 & 60.82
& 66.36 & 53.72
& 70.89 & 57.24
& 54.50 & 53.28 & 50.63
& 65.71 & 67.96
& 83.29 & 82.16
& 64.13 \\

Log-Rank
& Zero-shot
& 64.43 & 57.53
& 63.75 & 54.18
& 68.52 & 55.15
& 55.10 & 52.78 & 51.28
& 57.44 & 59.74
& 88.46 & 83.85
& 62.48 \\

LRR
& Zero-shot
& 65.47 & 55.45
& 64.93 & 53.01
& 68.53 & 57.99
& 54.61 & 52.73 & 57.41
& 57.09 & 58.15
& 85.99 & 80.56
& 62.46 \\

Log-Likelihood
& Zero-shot
& 63.71 & 56.36
& 62.97 & 53.13
& 67.97 & 54.38
& 53.37 & 51.77 & 50.73
& 57.92 & 59.28
& 88.48 & 83.75
& 61.83 \\

DNA-GPT
& Zero-shot
& 64.92 & 55.83
& 64.36 & 51.09
& 68.36 & 53.36
& 51.51 & 47.09 & 41.98
& 57.63 & 62.43
& 87.80 & 82.77
& 60.70 \\

Fast-DetectGPT
& Zero-shot
& 58.52 & 48.07
& 59.58 & 46.55
& 60.70 & 50.63
& 48.35 & 36.56 & 49.47
& 61.31 & 55.08
& 76.03 & 68.47
& 55.33 \\

Rank
& Zero-shot
& 51.34 & 44.97
& 50.33 & 42.06
& 57.08 & 48.83
& 42.61 & 41.49 & 38.84
& 41.67 & 46.65
& 83.86 & 80.00
& 51.52 \\

NPR
& Zero-shot
& 48.37 & 41.41
& 47.27 & 40.04
& 53.49 & 45.22
& 38.58 & 38.83 & 36.10
& 37.60 & 42.17
& 80.03 & 75.98
& 48.08 \\

DetectGPT
& Zero-shot
& 34.43 & 21.52
& 34.93 & 14.80
& 36.19 & 19.15
& 11.54 & 13.11 & 11.84
& 35.78 & 34.69
& 60.86 & 48.76
& 29.05 \\

Entropy
& Zero-shot
& 46.02 & 27.40
& 46.97 & 34.25
& 43.75 & 24.69
& 25.06 & 31.07 & 16.53
& 13.38 & 15.99
& 22.39 & 16.60
& 28.01 \\

\bottomrule
\end{tabular}%
}
\caption{Performance on the DetectRL benchmark. All values are reported as percentages.}
\label{tab:detectrl_leaderboard}

\medskip
\begin{minipage}{0.96\textwidth}
\footnotesize
\textit{Note:} Baseline results are taken from the original DetectRL
leaderboard. \pfour, Pangram 3.3.2, and Open Pangram are evaluated in a
\textbf{zero-shot} setting without benchmark-specific training and remain
competitive with or outperform the supervised methods. The average F1 is
computed over the test datasets.

\end{minipage}
\end{table}
}

\newcommand{\appendixOpAIBenchResults}{%
\begin{table}[H]
\centering
\small
\setlength{\tabcolsep}{4pt}
\renewcommand{\arraystretch}{1.08}
\resizebox{\columnwidth}{!}{%
\begin{tabular}{clccccc}
\toprule
Version
& Operation
& \begin{tabular}[c]{@{}c@{}}Actual AI character\\fraction (mean)\end{tabular}
& \begin{tabular}[c]{@{}c@{}}\pfour \\ Prediction (mean)\end{tabular}
& \begin{tabular}[c]{@{}c@{}}\pfour \\ Prediction (median)\end{tabular}
& \begin{tabular}[c]{@{}c@{}}Pangram 3.3.2\\edit score (mean)\end{tabular}
& \begin{tabular}[c]{@{}c@{}}Pangram 3.3.2\\edit score (median)\end{tabular} \\
\midrule
$v0$ & None          & 0.0000 & 0.0000 & 0.0000 & 0.0026 & 0.0021 \\
$v1$ & Polish        & 0.2075 & 0.0001 & 0.0000 & 0.0041 & 0.0022 \\
$v2$ & Paraphrase    & 0.3095 & 0.0318 & 0.0000 & 0.0184 & 0.0030 \\
$v3$ & Style rewrite & 0.4647 & 0.1359 & 0.0000 & 0.0460 & 0.0139 \\
$v4$ & Compress      & 0.4961 & 0.0548 & 0.0000 & 0.0370 & 0.0089 \\
$v5$ & Expand        & 0.6858 & 0.3257 & 0.2210 & 0.1238 & 0.0603 \\
$v6$ & Style rewrite & 0.8162 & 0.5617 & 0.6751 & 0.2073 & 0.1241 \\
$v7$ & Paraphrase    & 0.9257 & 0.6794 & 0.8415 & 0.2291 & 0.1359 \\
$v8$ & Polish        & 0.9935 & 0.6950 & 0.8853 & 0.2507 & 0.1444 \\
\bottomrule
\end{tabular}%
}
\caption{\pfour and Pangram 3.3.2 predictions across progressive
human--AI editing stages in the OpAI-Bench test set.}
\label{tab:opai-progressive-edits}

\medskip
\begin{minipage}{0.96\columnwidth}
\footnotesize
\textit{Note:} Each version contains 4,754 documents and represents a
successive stage in a cumulative human--AI editing trajectory.
The actual AI character fraction is provided by the character-level
ground-truth annotations in OpAI-Bench. For \pfour, the predicted
AI+Assisted fraction is calculated as
$\mathrm{fraction}_{\mathrm{AI}}+
\mathrm{fraction}_{\mathrm{AI\text{-}assisted}}$.
The mean and median
are calculated independently across the 4,754 documents at each stage.
\end{minipage}
\end{table}
}

\newcommand{\appendixSahaResults}{%
\begin{table}[H]
\centering
\small
\setlength{\tabcolsep}{4pt}
\renewcommand{\arraystretch}{1.08}
\resizebox{\textwidth}{!}{%
\begin{tabular}{l cccc cccc c}
\toprule
& \multicolumn{4}{c}{\textbf{Easy Subset}}
& \multicolumn{4}{c}{\textbf{Hard Subset}}
& \textbf{Human} \\
\cmidrule(lr){2-5}
\cmidrule(lr){6-9}
\textbf{Detector}
& \multicolumn{3}{c}{\textbf{TPR}}
& \textbf{FPR}
& \multicolumn{3}{c}{\textbf{TPR}}
& \textbf{FPR}
& \textbf{FPR} \\
\cmidrule(lr){2-4}
\cmidrule(lr){6-8}
& \textbf{AI-BP}
& \textbf{AI-EP}
& \textbf{AI-HI}
& \textbf{H-AI}
& \textbf{AI-BP}
& \textbf{AI-EP}
& \textbf{AI-HI}
& \textbf{H-AI}
& \textbf{H} \\
\midrule

\textbf{\pfour}
& 98.2
& \textbf{100.0}
& 96.3
& 1.4
& \textbf{100.0}
& \textbf{100.0}
& 98.9
& 2.5
& \textbf{0.0} \\

\textbf{Pangram 3.3.2}
& \textbf{100.0}
& \textbf{100.0}
& \textbf{100.0}
& 14.9
& \textbf{100.0}
& \textbf{100.0}
& \textbf{100.0}
& 4.5
& \textbf{0.0} \\
\midrule

Pangram 3.0
& \textbf{100.0}
& \textbf{100.0}
& \textbf{100.0}
& 3.0
& 97.0
& 99.3
& 92.6
& 3.1
& \textbf{0.0} \\

GPTZero
& 96.7
& 93.3
& 91.0
& 3.0
& 96.0
& 95.8
& 89.4
& 3.4
& 1.0 \\

LogLikelihood
& 97.8
& 96.4
& 72.4
& 0.4
& 46.0
& 40.9
& 30.5
& \textbf{0.0}
& \textbf{0.0} \\

LogLikelihood + Context
& 98.9
& 99.6
& 91.6
& 1.6
& 59.8
& 50.5
& 43.7
& 4.2
& 0.4 \\

Fast-DetectGPT
& \textbf{100.0}
& \textbf{100.0}
& 97.5
& 3.5
& 72.1
& 68.2
& 63.1
& 4.6
& 0.2 \\

Fast-DetectGPT + Context
& \textbf{100.0}
& \textbf{100.0}
& 99.3
& 9.0
& 75.3
& 73.9
& 68.2
& 9.5
& 0.9 \\

Binoculars
& 51.1
& 50.7
& 39.6
& \textbf{0.0}
& 21.6
& 22.3
& 14.1
& 0.1
& \textbf{0.0} \\

Binoculars + Context
& 50.7
& 53.3
& 33.7
& \textbf{0.0}
& 20.5
& 20.0
& 14.3
& 0.2
& \textbf{0.0} \\

\bottomrule
\end{tabular}%
}
\caption{Performance on the Saha peer-review benchmark.
All values are percentages.}
\label{tab:saha_peer_review}

\medskip
\begin{minipage}{0.96\textwidth}
\footnotesize
\textit{Note:} Baseline results are taken from Table~2 of the original
paper. For Pangram 3.3.2 and \pfour, AI-BP, AI-EP, and AI-HI are
successfully detected when the prediction is either AI or Mixed. For
H-AI, only an AI prediction is counted as a false positive; Mixed and
Human predictions are accepted. For fully human reviews (H), both AI
and Mixed predictions are counted as false positives. Pangram results
are evaluated on the test partition of each subset.
\end{minipage}
\end{table}
}

\newcommand{\appendixSemDetectResults}{%
\begin{table}[H]
\centering
\small
\setlength{\tabcolsep}{7pt}
\renewcommand{\arraystretch}{1.08}
\begin{tabular}{lccc}
\toprule
\textbf{Detector}
& \textbf{AUC (\%) $\uparrow$}
& \textbf{TPR@0.1\%FPR (\%) $\uparrow$}
& \textbf{TPR@1\%FPR (\%) $\uparrow$} \\
\midrule

\textbf{\pfour}
& $97.8 \pm 1.0$
& $95.5 \pm 1.0$
& $95.5 \pm 1.0$ \\

Pangram 3.3.2
& $\mathbf{100.0 \pm 0.0}$
& $\mathbf{100.0 \pm 0.0}$
& $\mathbf{100.0 \pm 0.0}$ \\

\midrule

LogRank
& $57.6 \pm 1.0$
& $0.0 \pm 0.0$
& $0.1 \pm 0.0$ \\

MAGE
& $69.9 \pm 1.0$
& $0.0 \pm 0.0$
& $0.8 \pm 1.0$ \\

Fast-DetectGPT
& $69.9 \pm 1.0$
& $2.1 \pm 1.0$
& $6.2 \pm 1.0$ \\

Binoculars
& $75.1 \pm 1.0$
& $0.8 \pm 1.0$
& $6.2 \pm 2.0$ \\

TF Model$^{\dagger}$
& $92.6 \pm 1.0$
& $36.9 \pm 6.0$
& $55.3 \pm 4.0$ \\

RADAR
& $96.5 \pm 0.0$
& $15.3 \pm 5.0$
& $37.1 \pm 6.0$ \\

Anchor$^{\dagger}$
& $97.9 \pm 0.0$
& $54.1 \pm 4.0$
& $71.3 \pm 7.0$ \\

EditLens$^{\dagger}$
& $\underline{99.8 \pm 0.0}$
& $60.6 \pm 16.0$
& $\underline{95.6 \pm 2.0}$ \\

Sem-Detect$^{\dagger}$
& $99.9 \pm 0.0$
& $\underline{76.0 \pm 11.0}$
& $97.3 \pm 3.0$ \\

\bottomrule
\end{tabular}
\caption{Performance on the Sem-Detect peer-review
benchmark. AUC and TPR values are reported as percentages.}
\label{tab:semdetect_binary}

\medskip
\begin{minipage}{0.96\columnwidth}
\footnotesize
\textit{Note:} Baseline results are taken from Table~1 of the original
paper. \pfour uses the derived AI likelihood, calculated as
$\mathrm{fraction}_{\mathrm{AI}}
+0.5\left(\mathrm{fraction}_{\mathrm{AI\text{-}assisted}}\right)$.
Uncertainty is estimated using 1,000 bootstrap resamples. All uncertainty
terms are reported in percentage points. Best and second-best results
are shown in bold and underlined, respectively. Detectors marked with
\textsuperscript{\textdagger} are domain-specific models trained or tuned on peer-review data.
\end{minipage}
\end{table}
}

\newcommand{\appendixEpochResults}{%
\begin{table}[H]
\centering
\begin{tabular}{lccc}
\toprule
Detector & Document-level FNR & Span-level FNR & FPR \\
\midrule
\pfour      & \textbf{2.86\%} (17/594) & \textbf{1.44\%} & 0.00\% (0/495)  \\
Pangram 3.3.2  & 5.05\% (30/594)          & 4.28\%          & 0.00\% (0/495)  \\
GPTZero        & 5.72\% (34/594)          & ---             & 0.00\% (0/495)  \\
Originality.ai & 9.09\% (54/594)          & ---             & 3.84\% (19/495) \\
\bottomrule
\end{tabular}
\caption{Epoch AI Style Imitation benchmark.}
\label{tab:detector-error-rates}

\medskip
\begin{minipage}{0.9\linewidth}
\footnotesize
\emph{Strict verdicts:} For document-level results, only a clean ``AI''
verdict counts as a detection; ``Mixed'' and ``AI-Assisted'' verdicts count as
non-detections (and as false positives on human text). Span-level results count
any non-AI span as a false negative in proportion to its length.
\end{minipage}
\end{table}
}

\subsection{Adversarial Robustness}
\label{subsec:adversarial-robustness}

We evaluate the robustness of \pfour to both humanizers and broader red-teaming attacks.

\subsubsection{Robustness to Humanizer Attacks}
We focus our evaluation of humanization on three domains in particular. These domains are a manually collected set of commercial humanizers chosen by top web traffic in early 2026; a popular GitHub repository containing instructions for AI agents to humanize their text; and an academic benchmark of AI texts attacked in specific manners.

\paragraph{How well does Pangram do at detecting humanized text?}
We find Pangram is robust to humanization attempts. We detect humanized text as AI-generated 97.67\% of the time and as either Mixed or AI-generated 98.83\% of the time.

\paragraph{How well does Pangram distinguish between AI-generated and humanized text?}
We evaluate the humanizer classifier on paired unmodified and humanized AI-generated texts. For this evaluation, humanized AI is the positive class and the corresponding unmodified AI text is the negative class. The humanizer classifier is only run if AI use is detected. \autoref{tab:humanizer-head-overall} reports aggregate performance. For this separate AI-versus-humanized-AI evaluation, the auxiliary head's per-system TPR ranges from 91.52\% to 99.39\% across commercial humanizers, as detailed alongside the primary-detector results in \autoref{tab:humanizer-eval}.

\begin{table}[htpb]
  \centering
  \scriptsize
  \setlength{\tabcolsep}{4pt}
  \begin{tabular}{@{}lrrrrrr@{}}
    \toprule
      & Accuracy & F1 & Precision
      & Recall & FPR & FNR \\
    \midrule
      & 96.82\% & 0.9678 & 98.02\% & 95.57\% & 1.93\% & 4.43\% \\
    \bottomrule
  \end{tabular}
  \caption{Auxiliary humanizer-head performance on paired unmodified and humanized AI-generated texts. A false negative is a humanized text predicted as AI-generated. A false positive is AI-generated text predicted as humanized.}
  \label{tab:humanizer-head-overall}
\end{table}

\begin{table}[t]
  \centering
  \small
  \setlength{\tabcolsep}{8pt}
  \begin{tabular}{@{}lrrr@{}}
    \toprule
    & \multicolumn{2}{c}{\pfour on humanized AI text}
    & \multicolumn{1}{c}{Auxiliary head} \\
    \cmidrule(lr){2-3}\cmidrule(l){4-4}
    Humanizer & AI recall & AI or Mixed recall & TPR \\
    \midrule
    Commercial A & 98.49\% & 99.40\% & 95.47\% \\
    Commercial B & 97.57\% & 99.09\% & 91.79\% \\
    Commercial C & 98.78\% & 98.78\% & 98.78\% \\
    Commercial D & 98.48\% & 99.39\% & 94.83\% \\
    Commercial E & 97.84\% & 98.77\% & 97.84\% \\
    Commercial F & 99.70\% & 100.00\% & 91.52\% \\
    Commercial G & 99.09\% & 99.39\% & 99.39\% \\
    Commercial H & 92.78\% & 95.28\% & 92.22\% \\
    Commercial I & 96.96\% & 98.78\% & 96.66\% \\
    Commercial J & 97.32\% & 98.66\% & 95.30\% \\
    Commercial K & 98.78\% & 99.39\% & 98.48\% \\
    Commercial L & 99.33\% & 100.00\% & 92.00\% \\
    Commercial M & 95.87\% & 98.73\% & 96.51\% \\
    \midrule
    \textbf{Overall commercial} & \textbf{97.69\%} & \textbf{98.83\%} & \textbf{95.57\%} \\
    \bottomrule
  \end{tabular}
  \caption{
 Performance of \pfour against different commercial evasion systems. The first two columns report TPRs for humanized text from the Pangram AI-use detector. The rightmost column is the auxiliary humanizer head's true-positive rate.}
  \label{tab:humanizer-eval}
\end{table}

\paragraph{BLADER}

BLADER\footnote{\href{https://github.com/blader/humanizer}
{\texttt{blader/humanizer} on GitHub}} is a popular GitHub repository that
contains instructions for LLMs on how to ``remove signs of AI-generated
writing from text.'' As of publication, BLADER has nearly 32,000 stars on
GitHub. We report our false negative rate on text humanized using three
frontier models prompted with BLADER's instructions in \autoref{tab:humanized-false negatives}

\begin{table}[ht]
\centering
\begin{tabular}{lrr}
\toprule
\textbf{Humanization model} &
\textbf{Documents} &
\textbf{False negatives, n (\%)} \\
\midrule
GPT-5.5    & 3,404  & 17 (0.499\%) \\
Opus 4.8   & 3,412  & 14 (0.410\%) \\
Sonnet 4.6 & 3,407  & 13 (0.382\%) \\
\midrule
\textbf{Overall} &
\textbf{10,223} &
\textbf{44 (0.430\%)} \\
\bottomrule
\end{tabular}
\caption{Performance of \pfour on text generated then rewritten using the BLADER series of agent skills. False negative rates on BLADER-adjusted AI-generated text, grouped by the model used for humanization.}
\label{tab:humanized-false negatives}
\end{table}

\newcommand{\appendixMELDEvalResults}{%
\begin{table}[H]
\centering
\small
\setlength{\tabcolsep}{5pt}
\renewcommand{\arraystretch}{1.08}
\resizebox{\textwidth}{!}{%
\begin{tabular}{lccccc}
\toprule
\textbf{Detector}
& \textbf{GPT-5.4-Mini}
& \textbf{Gemini-3-Flash}
& \textbf{Claude-Haiku-4.5}
& \textbf{Qwen-3.6-Plus}
& \textbf{Overall} \\
\midrule

\multicolumn{6}{l}{\textit{Zero-shot detectors}} \\

\textbf{\pfour}
& $\underline{99.989 \pm 0.008}$
& $\mathbf{99.991 \pm 0.008}$
& $\underline{99.989 \pm 0.009}$
& $\mathbf{99.993 \pm 0.006}$
& $\mathbf{99.990 \pm 0.004}$ \\

Pangram 3.3.2
& $99.962 \pm 0.015$
& $\underline{99.931 \pm 0.022}$
& $99.949 \pm 0.019$
& $\underline{99.985 \pm 0.010}$
& $\underline{99.957 \pm 0.009}$ \\

Open Pangram
& $57.2 \pm 1.1$
& $59.4 \pm 1.0$
& $65.1 \pm 1.1$
& $65.0 \pm 1.2$
& $61.7 \pm 1.0$ \\

GLTR
& $1.2 \pm 0.3$
& $0.5 \pm 0.3$
& $1.3 \pm 0.4$
& $0.3 \pm 0.1$
& $0.8 \pm 0.3$ \\

Fast-DetectGPT
& $9.2 \pm 1.7$
& $19.8 \pm 2.1$
& $14.3 \pm 2.0$
& $24.7 \pm 2.3$
& $17.0 \pm 2.0$ \\

Binoculars
& $0.2 \pm 0.1$
& $0.6 \pm 0.1$
& $1.1 \pm 0.2$
& $0.7 \pm 0.2$
& $0.6 \pm 0.1$ \\

\midrule
\multicolumn{6}{l}{\textit{Supervised detector baselines}} \\

RoBERTa-Large (OpenAI)
& $1.1 \pm 0.1$
& $1.1 \pm 0.1$
& $1.8 \pm 0.1$
& $1.8 \pm 0.1$
& $1.4 \pm 0.0$ \\

RoBERTa-ChatGPT
& $0.5 \pm 0.1$
& $1.2 \pm 0.2$
& $0.5 \pm 0.2$
& $0.9 \pm 0.2$
& $0.8 \pm 0.2$ \\

RADAR
& $0.4 \pm 0.2$
& $1.4 \pm 0.5$
& $2.2 \pm 0.8$
& $1.2 \pm 0.5$
& $1.3 \pm 0.5$ \\

Human-as-Outlier
& $2.7 \pm 1.2$
& $1.6 \pm 0.9$
& $0.8 \pm 0.4$
& $1.1 \pm 0.8$
& $1.6 \pm 0.8$ \\

ModernBERT-Detect
& $97.2 \pm 1.0$
& $93.0 \pm 2.2$
& $98.2 \pm 0.9$
& $93.7 \pm 2.0$
& $95.5 \pm 1.5$ \\

RepreGuard
& $23.0 \pm 1.3$
& $27.2 \pm 1.4$
& $41.2 \pm 1.8$
& $45.4 \pm 2.0$
& $34.2 \pm 1.6$ \\

MELD
& $\mathbf{100.0 \pm 0.0}$
& $99.7 \pm 0.1$
& $\mathbf{100.0 \pm 0.0}$
& $99.9 \pm 0.0$
& $99.9 \pm 0.0$ \\

\bottomrule
\end{tabular}%
}
\caption{TPR@1\%FPR ($\times 100$) on MELD-eval by generator.}
\label{tab:meld_eval_by_generator}

\medskip
\begin{minipage}{0.96\textwidth}
\footnotesize
\textit{Note:} Results for all detectors other than \pfour,
Pangram 3.3.2, and Open Pangram are taken from Table~4 of the
MELD paper. \pfour uses the derived AI likelihood
$\mathrm{fraction}_{\mathrm{AI}}+
0.5\left(\mathrm{fraction}_{\mathrm{AI\text{-}assisted}}\right).$
Confidence intervals for both Pangram models are 95\% percentile-bootstrap
intervals based on 5,000 resamples. The best and second-best results in
each column are shown in bold and underlined, respectively.
\end{minipage}
\end{table}

\begin{table}[H]
\centering
\small
\setlength{\tabcolsep}{4pt}
\renewcommand{\arraystretch}{1.08}
\caption{\pfour and Pangram 3.3.2 performance on MELD-eval by
attack type. AUROC and TPR values are reported as percentages.}
\label{tab:meld_eval_by_attack}
\resizebox{\textwidth}{!}{%
\begin{tabular}{llrrrrr}
\toprule
\textbf{Attack Type}
& \textbf{Detector}
& \textbf{AI $N$}
& \textbf{AUROC}
& \textbf{TPR@1\%FPR}
& \textbf{Detected AI}
& \textbf{False Negatives} \\
\midrule

None (clean)
& Pangram 3.3.2
& 31,448 & 99.99\% & 99.99\%
& 31,447 / 31,448 & 1 \\

None (clean)
& \pfour
& 31,448 & 100.00\% & 100.00\%
& 31,448 / 31,448 & 0 \\

\midrule

Homoglyph
& Pangram 3.3.2
& 31,448 & 99.99\% & 99.99\%
& 31,446 / 31,448 & 2 \\

Homoglyph
& \pfour
& 31,448 & 99.98\% & 99.97\%
& 31,440 / 31,448 & 8 \\

\midrule

Number perturbation
& Pangram 3.3.2
& 31,448 & 99.99\% & 99.98\%
& 31,444 / 31,448 & 4 \\

Number perturbation
& \pfour
& 31,448 & 100.00\% & 100.00\%
& 31,448 / 31,448 & 0 \\

\midrule

Synonym
& Pangram 3.3.2
& 31,448 & 99.98\% & 99.72\%
& 31,363 / 31,448 & 85 \\

Synonym
& \pfour
& 31,448 & 99.98\% & 99.96\%
& 31,438 / 31,448 & 10 \\

\midrule

Upper--lower flip
& Pangram 3.3.2
& 31,448 & 99.99\% & 99.99\%
& 31,447 / 31,448 & 1 \\

Upper--lower flip
& \pfour
& 31,448 & 99.99\% & 99.99\%
& 31,447 / 31,448 & 1 \\

\midrule

Whitespace
& Pangram 3.3.2
& 31,448 & 99.99\% & 99.99\%
& 31,447 / 31,448 & 1 \\

Whitespace
& \pfour
& 31,448 & 99.99\% & 99.99\%
& 31,446 / 31,448 & 2 \\

\midrule

Zero-width space
& Pangram 3.3.2
& 31,448 & 99.99\% & 99.99\%
& 31,447 / 31,448 & 1 \\

Zero-width space
& \pfour
& 31,448 & 100.00\% & 100.00\%
& 31,448 / 31,448 & 0 \\

\midrule

Overall
& Pangram 3.3.2
& 220,136
& 99.99\%
& 99.95\%
& 220,041 / 220,136
& 95 \\

\textbf{Overall}
& \pfour
& \textbf{220,136}
& \textbf{99.99\%}
& \textbf{99.99\%}
& \textbf{220,115 / 220,136}
& \textbf{21} \\

\bottomrule
\end{tabular}%
}

\medskip
\begin{minipage}{0.96\textwidth}
\footnotesize
\textit{Note:} Each attack is evaluated against the same 7,862 clean
human controls. Thresholds are calibrated independently at a target
FPR of 1\%. Attacks are applied only to AI texts.
\end{minipage}
\end{table}
}

\subsubsection{Red-Teaming}

\paragraph{Manual Red-Teaming} One week before launching \pfour, we gave all Pangram employees and a small group of early testers access to Pangram to stress-test it for false positives and false negatives. We used the feedback from the early access group to refine our calibration and postprocessing procedure and make small final adjustments to the model.

\paragraph{Automated Red-Teaming with AI Agents} We also gave two Codex agents with GPT-5.6 Sol access to the \pfour API for 24 hours and put them on ``Goal'' mode with Full Access to attempt to fool \pfour.
One agent was tasked with finding false positive gaps in our evaluation: kinds of text that are not explicitly present in our evaluation sets that could systematically have an elevated false positive rate. The false positive search covered a wide variety of historical and contemporary literature, official and legal records, patents and filings, public minutes, engineering and incident records, medical reports, field notes, learner writing, and interviews. Zero new false positives were found by the false positive red-teaming agent in 24 hours.

The second agent was tasked with finding systematic bypasses that could create false negatives repeatedly. This agent tested several potential bypass hypotheses, such as imitating specific human voices and styles, using different operational and professional registers, imitating a notetaker, using much terser vocabulary, conditioning its output on detailed source writing, aggregating multiple AI documents together, reformatting data as PDFs, putting the text through multiple rounds of translation, and more. The only bypass explicitly found by the agent was acting as a surgeon/pathologist dictating notes aloud. We examined these false negatives by eye and deemed them to be out of scope: more context was given to the LLM than the output, and the outputs more resembled extremely terse factual bullets rather than open-ended prose.

From these two experiments, we deemed the model resilient to agent-based adversarial attacks and sufficiently robust to release to the public.

%% file: sections/5_public_benchmark_summary.tex
\newcommand{\publicmetriccheck}{%
  \textcolor{pangramforest}{\ensuremath{\checkmark}}}
\newcommand{\publicmetricnone}{\textcolor{pangramgray}{---}}

\begin{table}[H]
\centering
\scriptsize
\setlength{\tabcolsep}{4.2pt}
\renewcommand{\arraystretch}{1.14}
\begin{tabular}{@{}lcccccc@{}}
\toprule
Benchmark
& \shortstack{AUROC /\\AUC}
& \shortstack{TPR at\\fixed FPR}
& FPR
& FNR
& \shortstack{Mean\\accuracy}
& F1
\\
\midrule
\multicolumn{7}{@{}l}{\textbf{\color{pangrambrown}Binary}} \\
\addlinespace[0.1em]
\hspace{0.6em}UChicago \cite{jabarian2025artificial}
  & \publicmetriccheck & \publicmetriccheck & \publicmetriccheck
  & \publicmetricnone & \publicmetricnone & \publicmetricnone
  \\
\hspace{0.6em}VUB~\cite{vanvlasselaer2026whowrote}
  & \publicmetricnone & \publicmetricnone & \publicmetricnone
  & \publicmetricnone & \publicmetricnone & \publicmetricnone
  \\
\hspace{0.6em}GEDE~\cite{gehring2025gede}
  & \publicmetriccheck & \publicmetriccheck & \publicmetricnone
  & \publicmetricnone & \publicmetricnone & \publicmetricnone
  \\
\hspace{0.6em}Perkins et al.~\cite{perkins2024simple}
  & \publicmetricnone & \publicmetricnone & \publicmetricnone
  & \publicmetricnone & \publicmetriccheck & \publicmetricnone
  \\
\hspace{0.6em}Epoch AI Style Imitation~\cite{epoch2026aidetectorsfalsenegatives}
  & \publicmetricnone & \publicmetricnone & \publicmetriccheck
  & \publicmetriccheck & \publicmetricnone & \publicmetricnone
  \\
\hspace{0.6em}DetectRL~\cite{wu2024detectrl}
  & \publicmetriccheck & \publicmetricnone & \publicmetricnone
  & \publicmetricnone & \publicmetricnone & \publicmetriccheck
  \\
\hspace{0.6em}MELD-eval~\cite{li2026meld}
  & \publicmetriccheck & \publicmetriccheck & \publicmetricnone
  & \publicmetricnone & \publicmetricnone & \publicmetricnone
  \\
\hspace{0.6em}Sem-Detect~\cite{duarte2026semdetect}
  & \publicmetriccheck & \publicmetriccheck & \publicmetricnone
  & \publicmetricnone & \publicmetricnone & \publicmetricnone
  \\
\addlinespace[0.35em]
\multicolumn{7}{@{}l}{\textbf{\color{pangrambrown}Mixed}} \\
\addlinespace[0.1em]
\hspace{0.6em}Saha Peer Review~\cite{saha2026reviewpolishing}
  & \publicmetricnone & \publicmetriccheck & \publicmetriccheck
  & \publicmetricnone & \publicmetricnone & \publicmetricnone
  \\
\hspace{0.6em}OpAI-Bench~\cite{bsharat2026operationguidedprogressivehumantoaitext}
  & \publicmetricnone & \publicmetricnone & \publicmetricnone
  & \publicmetricnone & \publicmetricnone & \publicmetricnone
  \\
\bottomrule
\end{tabular}
\caption{Metric coverage across the binary and mixed-authorship evaluations.
A checkmark means that Section~5 reports that metric for \pfour.}
\label{tab:public-benchmark-metric-map}
\end{table}

\paragraph{Summary of results}
\autoref{tab:public-benchmark-master} compares \pfour, Pangram 3, and the strongest non-Pangram comparator on every split reported in the detailed benchmark tables. Comparisons between benchmarks are challenging, as there is no unified metric of success; some choose FNR/FPR, while others look at AUROC or other locally defined metrics. A summary of what metrics are presented in each benchmark is shown in \autoref{tab:public-benchmark-metric-map}.

\begingroup
\scriptsize
\setlength{\tabcolsep}{3pt}
\setlength{\LTleft}{0pt}
\setlength{\LTright}{0pt}
\renewcommand{\arraystretch}{1.12}
\begin{longtable}{@{}
  >{\raggedright\arraybackslash}p{0.13\linewidth}
  >{\raggedright\arraybackslash}p{0.17\linewidth}
  >{\raggedright\arraybackslash}p{0.27\linewidth}
  >{\raggedright\arraybackslash}p{0.16\linewidth}
  >{\raggedright\arraybackslash}p{0.20\linewidth}
@{}}
\toprule
\textbf{Benchmark}
& \textbf{Evaluation split}
& \textbf{\pfour result}
& \textbf{Pangram 3 result}
& \textbf{Next-best result} \\
\midrule
\endfirsthead
\toprule
\textbf{Benchmark}
& \textbf{Evaluation split}
& \textbf{\pfour result}
& \textbf{Pangram 3 result}
& \textbf{Next-best result} \\
\midrule
\endhead
\midrule
\multicolumn{5}{r}{\textit{Continued on next page}} \\
\endfoot
\bottomrule
\endlastfoot

\multicolumn{5}{@{}l}{\textbf{\color{pangrambrown}Binary}} \\
\midrule
UChicago
& Standard, full length
& TPR@1\%FPR: 100.00\%
& 99.86\%
& GPTZero: 98.64\% \\
& Standard, under 50 words
& TPR@1\%FPR: 99.70\%
& 88.87\%
& GPTZero: 0.00\% \\
& Humanizer, full length
& TPR@1\%FPR:  98.93\%
& 98.08\%
& GPTZero: 44.32\% \\
& Humanizer, under 50 words
& TPR@1\%FPR:  73.32\%
& 61.74\%
& GPTZero: 0.00\% \\
& Human controls
& FPR: 0.00\%
& 0.00\%
& Originality: 0.11\% \\

\midrule
VUB
& Fully AI-generated
& Recall: 100\%
& 97.4\%
& GPTZero: 0\% \\

\midrule
GEDE
& Overall
& TPR@1\%FPR: 100.0\%
& 98.9\%
& GPTZero: 77.6\% \\
& Fully generated
& TPR@1\%FPR: 100.0\%
& 100.0\%
& GPTZero: 93.2\% \\
& AI-improved human
& TPR@1\%FPR: 100.0\%
& 97.4\%
& GPTZero: 54.7\% \\
& Humanized
& TPR@1\%FPR:  100.0\%
& 100.0\%
& GPTZero: 92.6\% \\

\midrule
Perkins et al.
& Baseline AI
& Mean accuracy: 100.0\%
& ---
& Copyleaks: 73.9\% \\
& Manipulated AI
& Mean accuracy: 94.1\%
& ---
& Copyleaks: 58.7\% \\

\midrule
Epoch AI Style Imitation
& Overall
& FNR / FPR: 2.86\% / 0.00\%
& 5.05\% /  0.00\%
& GPTZero: 5.72\% / 0.00\% \\

\midrule
DetectRL
& Multi-domain
& F1:  96.93
&  98.05
& Rob-Base: 99.75 \\
& Multi-LLM
& F1: 97.25
& 98.39
& Rob-Base: 99.58 \\
& Multi-attack
& F1: 96.00
& 97.53
& Rob-Large: 99.03 \\
& Domain generalization
& F1: 96.95
& 98.05
& Rob-Base: 83.00 \\
& LLM generalization
& F1: 97.25
& 98.39
& X-Rob-Base: 92.73 \\
& Attack generalization
& F1: 95.99
& 97.53
& Rob-Base: 92.37 \\
& Time, test
& F1: 91.30
& 93.73
& Rob-Large: 85.23 \\
& Human writing
& F1: 90.77
&  91.95
& Rob-Base: 97.34 / Rob-Large: 94.63 \\
& Overall
& Average F1: 95.30
& 96.70
& Rob-Base: 93.02 \\

\midrule
MELD-eval
& GPT-5.4-Mini
& TPR@1\%FPR: 99.989\%
& 99.962\%
& MELD: 100.0\% \\
& Gemini-3-Flash
& TPR@1\%FPR: 99.991\%
& 99.931\%
& MELD: 99.7\% \\
& Claude-Haiku-4.5
& TPR@1\%FPR: 99.989\%
& 99.949\%
& MELD: 100.0\% \\
& Qwen-3.6-Plus
& TPR@1\%FPR: 99.993\%
& 99.985\%
& MELD: 99.9\% \\
& Generator overall
& TPR@1\%FPR: 99.990\%
& 99.957\%
& MELD: 99.9\% \\
& No attack
& TPR@1\%FPR: 100.00\%
& 99.99\%
& --- \\
& Homoglyph
& TPR@1\%FPR: 99.97\%
& 99.99\%
& --- \\
& Number perturbation
& TPR@1\%FPR: 100.00\%
& 99.98\%
& --- \\
& Synonym
& TPR@1\%FPR: 99.96\%
& 99.72\%
& --- \\
& Upper--lower flip
& TPR@1\%FPR: 99.99\%
& 99.99\%
& --- \\
& Whitespace
& TPR@1\%FPR:99.99\%
& 99.99\%
& --- \\
& Zero-width space
& TPR@1\%FPR:  100.00\%
& 99.99\%
& --- \\
& Attack overall
& TPR@1\%FPR: 99.99\%
&  99.95\%
& --- \\

\midrule
Sem-Detect
& Binary detection
& TPR@0.1\%FPR :
  95.5\%
& 100.0\%
& Sem-Detect: 76.0\% \\

\midrule
\multicolumn{5}{@{}l}{\textbf{\color{pangrambrown}Mixed}} \\
\midrule
Saha Peer Review
& Easy, AI-BP
& TPR: 98.2\%
& 100.0\%
& Fast-DetectGPT: 100.0\% \\
& Easy, AI-EP
& TPR: 100.0\%
& 100.0\%
& Fast-DetectGPT: 100.0\% \\
& Easy, AI-HI
& TPR: 96.3\%
& 100.0\%
& Fast-DetectGPT + Context: 99.3\% \\
& Easy, H-AI
& FPR: 1.4\%
& 14.9\%
& Binoculars: 0.0\% \\
& Hard, AI-BP
& TPR: 100.0\%
& 100.0\%
& GPTZero: 96.0\% \\
& Hard, AI-EP
& TPR: 100.0\%
& 100.0\%
& GPTZero: 95.8\% \\
& Hard, AI-HI
& TPR: 98.9\%
& 100.0\%
& GPTZero: 89.4\% \\
& Hard, H-AI
& FPR: 2.5\%
& 4.5\%
& LogLikelihood: 0.0\% \\
& Human
& FPR: 0.0\%
& 0.0\%
& LogLikelihood: 0.0\% \\

\midrule
OpAI-Bench
& $v0$, none (target: 0.00\%)
& AI+Assisted mean: 0.00\% 
& 0.26\% 
& --- \\
& $v1$, polish (target: 20.75\%)
& AI+Assisted mean: 0.01\% 
& 0.41\% 
& --- \\
& $v2$, paraphrase (target: 30.95\%)
& AI+Assisted mean: 3.18\% 
& 1.84\% 
& --- \\
& $v3$, style rewrite (target: 46.47\%)
& AI+Assisted mean: 13.59\% 
& 4.60\% 
& --- \\
& $v4$, compress (target: 49.61\%)
& AI+Assisted mean: 5.48\% 
& 3.70\% 
& --- \\
& $v5$, expand (target: 68.58\%)
& AI+Assisted mean: 32.57\% 
& 12.38\% 
& --- \\
& $v6$, style rewrite (target: 81.62\%)
& AI+Assisted mean: 56.17\% 
& 20.73\%
& --- \\
& $v7$, paraphrase (target: 92.57\%)
& AI+Assisted mean: 67.94\% 
& 22.91\% 
& --- \\
& $v8$, polish (target: 99.35\%)
& AI+Assisted mean: 69.50\% 
& 25.07\% 
& --- \\
\end{longtable}
\captionof{table}{Results on the native metrics reported for each public benchmark. The
next-best entry is the strongest non-Pangram result available for each metric. Full comparisons appear in
\S\ref{app:detailed-benchmark-results}.}
\label{tab:public-benchmark-master}
\endgroup

%% file: sections/7_conclusion.tex
\section{Conclusion}
\label{sec:conclusion}

We introduce our most powerful AI detector model, \pfour, with the purpose of detecting more granular AI usage and increasing both recall and specificity over Pangram 3.3.2. \pfour is the first AI-detection model that can predict not only document-level AI usage, but also homogeneous and heterogeneous mixed text. We intend \pfour to increase visibility into how an author crafted a text, while still managing to keep our overall FPR at 0.0041\%.

\paragraph{Limitations} While \pfour is a state-of-the-art AI detection model, its estimates are statistical, and both false positives and false negatives, although rare, do occur. Its predictions are also to some degree a black box, and the reasons it makes particular predictions are difficult to understand. Predictions for the same text in different contexts may also be inconsistent. For example, a section of a paper run in isolation may receive a different prediction than it does within the full context of the paper. In such cases, the additional context may give the model more information with which to make a holistic decision. We are working to reduce these inconsistencies over time. Finally, \pfour does not account for people who intentionally write like LLMs or who have absorbed elements of LLM style into their writing. This form of data drift is a major focus of our ongoing research.

Looking forward, we look to answer questions related to data drift,
consistency, and interpretability, and to work toward the overall goal of
understanding the provenance of AI-generated text and how it influences the
world around us.

\section*{AI Disclosure}
\label{sec:disclosure}

We use AI agents extensively for coming up with research ideas, processing data, writing code, running experiments, and evaluating results. We use some AI assistance in the writing of the technical report, for proofreading and review, but the majority of the writing is ours. We claim full responsibility for the factual accuracy of the claims we make in this report. For full disclosure, we provide a link to the \pfour result on this technical report, and agree with its assessment of our authorship.\footnote{\url{https://www.pangram.com/history/6c43abfc-803e-4dd5-96d1-8eba2768ba39}}

\begin{figure}[h!]
  \centering
  \includegraphics[width=0.5\textwidth]{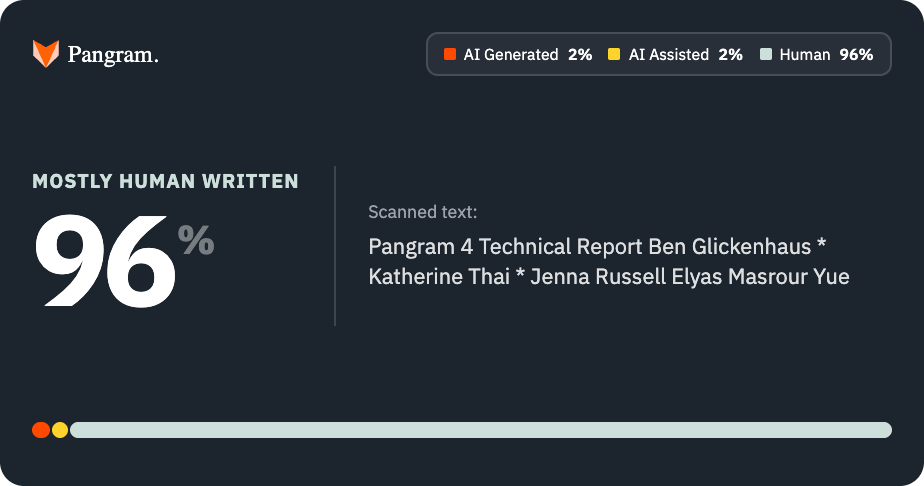}
  \caption{Pangram AI result for this technical report.}
  \label{fig:pangram-result}
\end{figure}

\section*{Acknowledgements}
\label{sec:acknowledgements}

We would like to thank the following people for their contributions to \pfour. We thank Artem Frenk for early discussions regarding tokenwise postprocessing and the HMM/CRF formulation. We thank Fanyi Pan, Lu Lyu, Anna Lin, and Tianxu Zhou for \pfour product support, design, UI/UX engineering, and testing to bring \pfour to our users. We thank Jocelyn James for graphic design contributions. We thank Annamieka Aerts and James Freedman for helping with the technical writing and proofreading. We thank Levi Goldstein and Lauryl Salami for data expertise and contributions. We also thank our early testers for providing invaluable pre-release feedback on the model.

%% file: sections/appendix.tex
\section{Citation reference}
\label{app:citation-reference}

\begin{infobox}[Citation] Please use the following BibTeX entry when citing this work:
\begin{quote}
\begin{tabular}{@{}l@{}}
\texttt{@techreport\{pangram2026pangram4,}\\
\texttt{\hspace*{1em}title=\{Pangram 4 Technical Report\},}\\
\texttt{\hspace*{1em}author=\{Ben Glickenhaus and Katherine Thai and}\\
\texttt{\hspace*{2em}Jenna Russell and Elyas Masrour and Yue Han and}\\
\texttt{\hspace*{2em}Max Spero and Bradley Emi\},}\\
\texttt{\hspace*{1em}institution=\{Pangram Labs\},}\\
\texttt{\hspace*{1em}year=\{2026\}}\\
\texttt{\}}
\end{tabular}
\end{quote}
\end{infobox}

\section{Mixed Authorship Text Dataset}
In tables \ref{app:polish-subset} and \ref{app:edits-subset}, we provide the prompts and instructions used to create our AI Polish and AI Editing Prompts evaluation datasets.

\begin{table*}[h]
\centering
\footnotesize
\renewcommand{\arraystretch}{1.15}
\begin{tabular}{
    @{}
    p{0.27\textwidth}
    p{0.68\textwidth}
    @{}
}
\toprule
\textbf{Consumer product / LLM}
& \textbf{Editing instruction / prompt} \\
\midrule

Apple Intelligence in Pages &
Proofread \\

\addlinespace

Grammarly
& Fix any mistakes \\

\addlinespace

Claude Opus 4.8 \newline
Gemini 3.1 Pro Preview \newline
GPT-5.5
&
Fix spelling, punctuation, and clear grammar errors only. \newline
Smooth awkward phrasing while preserving the author's sentence structure. \newline
Make the writing slightly clearer without changing tone or vocabulary level. \newline
Correct minor academic style issues while keeping the draft recognizably human. \newline
Lightly polish for readability. \newline
Make minimal copyediting changes suitable for a final proofreading pass. \newline
Change approximately one tenth of the text. \\

\bottomrule
\end{tabular}
\caption{Editing instructions and prompts used to create our AI Polish evaluation set.}
\label{app:polish-subset}
\end{table*}

\begin{table*}[h]
\centering
\footnotesize
\renewcommand{\arraystretch}{1.15}
\begin{tabular}{
    @{}
    p{0.27\textwidth}
    p{0.68\textwidth}
    @{}
}
\toprule
\textbf{Consumer product / LLM}
& \textbf{Editing instruction / prompt} \\
\midrule

Apple Intelligence
&
Make concise \newline
Rewrite \\

\addlinespace

Gemini in Google Docs
&
Rephrase \\

\addlinespace

Grammarly
&
Change writing style: Professional \newline
Change writing style: Academic \newline
Change writing style: Creative \newline
Sound fluent \newline
Improve it \newline
Simplify it \newline
Paraphrase it \\

\addlinespace

Claude Opus 4.8 \newline
Gemini 3.1 Pro Preview \newline
GPT-5.5
&
Substantially rewrite for polished academic style while preserving meaning. \newline
Restructure sentences and paragraphs for stronger argument flow. \newline
Substantially transform the draft into publication-ready academic prose. \newline
Heavily revise wording, organization, and transitions without changing facts. \newline
Improve logic, emphasis, and readability through extensive rewriting. \newline
Produce a highly polished version that remains faithful to the source. \newline
Change approximately half of the text. \newline
Change approximately three quarters of the text. \\

\bottomrule
\end{tabular}
\caption{Editing instructions and prompts used to create our AI Editing Prompts evaluation set.}
\label{app:edits-subset}
\end{table*}

\section{Detailed Benchmark Results}
\label{app:detailed-benchmark-results}

This section expands upon the results in \autoref{tab:public-benchmark-master}.

\subsection{Binary}

\appendixUChicagoResults

\appendixVUBResults

\appendixGEDEResults

\appendixPerkinsResults

\appendixEpochResults

\appendixDetectRLResults

\appendixMELDEvalResults

\appendixSemDetectResults

\subsection{Mixed}

\appendixSahaResults

\appendixOpAIBenchResults